\documentclass[oneside,12pt]{report}
\usepackage{listings}
\usepackage{lmodern}
\usepackage[T1]{fontenc}
\usepackage{booktabs}
\usepackage{amsfonts}
\usepackage[paper=a4paper]{geometry}
\usepackage{amsmath}
\usepackage{float}
\usepackage[utf8]{inputenc}
\usepackage{graphicx}
\usepackage{svg}
\usepackage{hyperref}
\usepackage{enumitem}
\hypersetup{
    colorlinks=false,
    breaklinks=true
}
\makeatletter
\let\orig@addtocontents\addtocontents
\makeatother

\usepackage[title, titletoc]{appendix}

\makeatletter
\let\addtocontents\orig@addtocontents
\makeatother

\usepackage[
  backend=biber,
  style=authoryear,
  sorting=nyt,    
  natbib=true,
  url=false,
  doi=true
]{biblatex}
\usepackage{datetime}
\newdateformat{monthyeardate}{%
  \monthname[\THEMONTH], \THEYEAR}

\newcommand\thesistitle{Diagnostics of cognitive failures in multi-agent expert systems using dynamic evaluation protocols and subsequent mutation of the processing context} 
\newcommand\authorname{Andrejs Sorstkins} 
\newcommand\authordegrees{BSc (Hons) Computer Science} 
\newcommand\supervisor{Dr Alistair Baron, Josh Bailey} 

\newgeometry{left=25mm, right=25mm, top=25mm, bottom=25mm, includeheadfoot}

\usepackage{fancyhdr}
\fancypagestyle{preamble}{ %
  \fancyhf{} 
  \fancyfoot[C]{\thepage}
  
}
\fancypagestyle{main}{
  \fancyhf{}
  \fancyhead[R]{\thepage}
  \fancyfoot[C]{\thepage}
  \fancyhead[L]{\textit{ \nouppercase{\leftmark}} }
  \fancyhead[R]{\textit{ \nouppercase{\rightmark}} }
  
}

\fancypagestyle{plain}{ %
  \fancyhf{} 
  \fancyfoot[C]{\thepage} 
  
}

\usepackage{lipsum} 

\begin{document}
\pagenumbering{roman}

\begin{titlepage}

\center

\huge  \textbf{\thesistitle}

\vfill

\includegraphics[width=0.8
\linewidth]{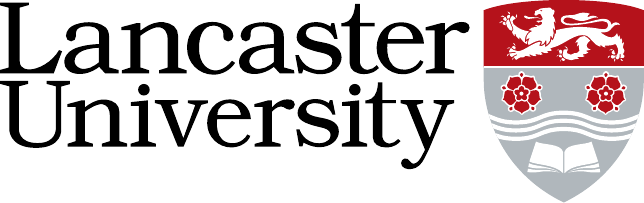}


\large \textbf{\authorname} \vspace{1mm}

\textbf{\authordegrees}

\vfill

A dissertation submitted for the degree of

\textit{Master of Science} in Data Science

\vfill

Supervised by \textit{\supervisor}

\vfill

School of Computing and Communications \vspace{1mm}

Lancaster University

\vfill

\monthyeardate\today

\end{titlepage}

\clearpage

\section*{\centering Declaration} \vspace{24pt}

I declare that the work presented in this dissertation is, to the best of my knowledge and belief, original and my own work. The material has not been submitted, either in whole or in part, for a degree at this, or any other university. \\
\

\noindent Name: \textbf{\authorname} \newline
\noindent Date: \textbf{\monthyeardate\today}

\clearpage

\begin{center}
\textbf{\thesistitle}

\authorname, \authordegrees.

School of Computing and Communications, Lancaster University

A dissertation submitted for the degree of \textit{Master of Science} in Data Science
\newline \monthyeardate\today
\end{center}

\section*{\centering Abstract} \vspace{12pt}

\textbf{The rapid evolution of neural architectures—from multilayer perceptrons to large-scale Transformer-based models—has enabled language models (LLMs) to exhibit emergent agentic behaviours when equipped with memory, planning, and external tool use. However, their inherent stochasticity and multi-step decision processes render classical evaluation methods inadequate for diagnosing agentic performance. This work introduces a diagnostic framework for expert systems that not only evaluates but also facilitates the transfer of expert behaviour into LLM-powered agents. The framework integrates (i) curated golden datasets of expert annotations, (ii) silver datasets generated through controlled behavioural mutation, and (iii) an LLM-based Agent Judge that scores and prescribes targeted improvements. These prescriptions are embedded into a vectorized recommendation map, allowing expert interventions to propagate as reusable improvement trajectories across multiple system instances. We demonstrate the framework on a multi-agent recruiter-assistant system, showing that it uncovers latent cognitive failures—such as biased phrasing, extraction drift, and tool misrouting—while simultaneously steering agents toward expert-level reasoning and style. The results establish a foundation for standardized, reproducible expert behaviour transfer in stochastic, tool-augmented LLM agents, moving beyond static evaluation to active expert system refinement.}

\clearpage

\clearpage

\section*{\centering Acknowledgements} \vspace{12pt}

I’d like to thank my academic supervisor, Dr Alistair Baron, and my
company supervisors Josh Bailey and Dr. Alice Ashcroft for guiding and mentoring me through all stages of this project’s completion. Sincere thanks to Dr Simon Tomilison for giving me the opportunity to work on this project. I’d also like to thank my colleagues for their invaluable insights
and discussions. Finally, I want to thank my parents and God for
always investing in my skills and enabling me to complete this
dissertation

\clearpage

\tableofcontents

\clearpage

\listoffigures

\clearpage

\listoftables

\clearpage
\pagestyle{main}
\pagenumbering{arabic}

\chapter{Introduction}
\section{Technological Evolution and Motivation}

Over the past decade, neural architectures have undergone a profound transformation, shifting from static classifiers to dynamic, context-aware generators capable of agentic reasoning. Early multilayer perceptrons (MLPs) captured nonlinear input–output relationships but lacked temporal memory or interpretability \citep{Rumelhart1986Backprop}. Recurrent neural networks (RNNs) and long short-term memory networks (LSTMs) partially addressed this by introducing sequence memory, yet their limited effective context and vanishing gradient problems constrained their utility \citep{Hochreiter1997LSTM}. The emergence of self-attention and the Transformer architecture \citep{Vaswani2017Attention} removed recurrence altogether, enabling each token to attend globally across the sequence. This paradigm shift, amplified through scaling \citep{Brown2020GPT3}, gave rise to large language models (LLMs) with strong zero- and few-shot generalization.

Building on these foundations, agentic frameworks such as ReAct \citep{Yao2022ReAct}, Toolformer \citep{Schick2023Toolformer}, and AutoGen \citep{Wu2023AutoGen} extended autoregressive models into multi-step systems capable of planning, tool use, and self-reflection (see also Reflexion \citep{Shinn2023Reflexion}). These architectures effectively transformed LLMs into stochastic agents operating within dynamic environments. However, this transition introduced new challenges: context drift, tool misuse, latent bias propagation, and reasoning incoherence—failure modes surfaced by recent tool-use and robustness evaluations. Classical evaluation paradigms—designed for deterministic models and static benchmarks—are ill-suited to capture these behaviors, as they presuppose a fixed input–output mapping \citep{Hendrycks2021MMLU,Srivastava2022BIGBench}.

Consequently, the field has begun to explore diagnostic methodologies that can both detect and remediate cognitive failures in agentic LLMs. Such diagnostics instrument intermediate reasoning states, decision trajectories, and tool invocations to surface errors invisible to end-task metrics 

. Yet evaluation alone is insufficient: for high-stakes applications, systems must not only be measured against expert standards but also be steered toward them. This dissertation addresses this gap by proposing a diagnostic framework for expert systems that operationalizes expert behaviour transfer. By curating expert demonstrations, mutating them into scalable silver datasets, and embedding Agent Judge prescriptions into a reusable recommendation graph, the framework transforms evaluation into an active mechanism of expert knowledge propagation. In doing so, it advances LLM evaluation from static performance reporting to dynamic, reproducible refinement toward expert-level competence.

\section{Company and Project Overview}

JobFair Software (trading as JobFair) is a boutique desktop‐computing software vendor specializing in AI‐driven recruitment tools that embed proprietary, peer‐reviewed gender‐decoder technology directly into ATS and HR workflows to eliminate biased language in job descriptions and enable recruiters to focus on selecting the best person. This dissertation evaluates JobFair’s emerging agentic recruiter‐assistant system, which augments that core decoder with autonomous planning and external tool integration by assessing its ability to interpret and rewrite live job descriptions while maintaining semantic fidelity and bias mitigation, measuring planning coherence across multi‐step workflows, and evaluating alignment with expert behavior. Aims of this dissertation is to make a diagnostic system which will be able to improve JobFair bias mitigation system.

\subsection*{Agent Diagnostics Objectives}

We propose that the core objective of Agent Diagnostics is twofold:

\begin{enumerate}[label=\arabic*., leftmargin=*]
  \item \textbf{Behavioral Mutation:}  
    Identify the characteristic features of expert agent judgments—lexical choices, argument structures, tool‐use patterns—and encode these into context mutations. This may involve inserting annotated exemplars, weighted attention priors, or synthetic “thought traces” that reflect expert reasoning.
  
  \item \textbf{Propagation Assessment:}  
    Systematically evaluate how these context mutations propagate through successive agent interactions. Key metrics include reduction in hallucination rate, convergence of answer quality toward an expert standard, and stability of tool‐invocation sequences under varied query loads.
\end{enumerate}

\section{Research Questions}

The dissertation investigates how to \emph{diagnose} and \emph{steer} expert LLM agents along two orthogonal axes, (i) sentence-level \textbf{ extraction diagnostic (ED)} over $E(x)$ and (ii) \textbf{ behavior diagnostic (BD)} over tone, style, and reasoning relative to expert thoughts $T(x)$ using a compact golden set $\mathcal{G}$, a mutated silver set and an LLM Agent Judge that both \emph{scores} and \emph{recommends} improvements.

\begin{enumerate}
  \item \textbf{Silver mutation feasibility.} 
  Can the Agent Mutator generate behavior-aligned (expert-style) silver instances without copying? 

  \item \textbf{Judge validity \& usefulness.} 
  Do the Agent Judge’s ED/BD scores and prescriptions align with expert annotations and audits on the golden set—i.e., are its recommendations correct and actionable for experts? 

  \item \textbf{Failure-mode discovery in GLS.} 
  To what extent do the ED and BD diagnostics surface cognitive failures in the production JobFair Gendered Language System (GLS)—e.g., over/under-extraction, stylistic drift, tool misrouting—that static single-pass metrics miss? 

\end{enumerate}

\chapter{Background}
\section{Early Neural Language Models: From MLPs to LSTMs}

LLM-based agent systems are stochastic and can behave unpredictably. To operate them reliably, understand the diagnostics introduced in Chapter 3, or devise new methodologies, the reader must have a clear view of the system’s components and how they interact. This chapter provides that foundation. The chapter establishes terminology, assumptions, and evaluation context for Chapter 3: Methodology, including the Agent Diagnostic Method for Expert Systems (ADM–ES), the Text Block Extraction Diagnostic (ED), the Behaviour Diagnostic (BD), and the agent Mutator with its acceptance checks (e.g., mean BERTScore over k). It thereby provides the conceptual baseline needed to interpret results and reproduce procedures in subsequent chapters.

The evolution of large language models (LLMs) is rooted in breakthroughs in neural network design and training strategies that have progressively enabled richer representations, longer context modeling, and more powerful generative capabilities. The journey begins with multilayer perceptrons (MLPs), which demonstrated that layered networks of simple nonlinear units could learn complex input–output mappings through the backpropagation algorithm. Although these early networks excelled at pattern recognition tasks, they operated on fixed-size inputs and lacked any mechanism to capture sequential dependencies or memory of previous inputs.\citep{Rumelhart1986Backprop}

To address sequence modeling, researchers developed recurrent neural networks (RNNs), which introduce a hidden state that updates dynamically as new inputs arrive. By maintaining and modifying an internal memory, RNNs could, in principle, remember information from earlier in a sequence. However, practical training of RNNs suffered from vanishing or exploding gradients, making it difficult to learn dependencies spanning more than a few time steps.

The Long Short-Term Memory (LSTM) architecture, introduced by \citep{Hochreiter1997LSTM}, overcame these training challenges by incorporating gating mechanisms—specifically, input, forget, and output gates—that regulate the flow of information into and out of the cell state. This design enabled robust learning of long-range dependencies, and LSTMs rapidly became the backbone of state-of-the-art systems in language modeling, machine translation, and other sequential tasks.

Despite these advances, RNN-based approaches still processed sequence elements in a largely linear, step-by-step fashion. The introduction of attention mechanisms by \citep{Bahdanau2015NMT} marked a pivotal shift: instead of compressing an entire sequence into a fixed-size vector, attention allows the model to compute a context-dependent weighted sum of all encoder hidden states at each decoding step. This dynamic focus mechanism led to significant improvements in translation quality and opened the door to architectures that rely less on recurrence.

\begin{figure}[h]
    \centering
    \includegraphics[width=0.4\textwidth]{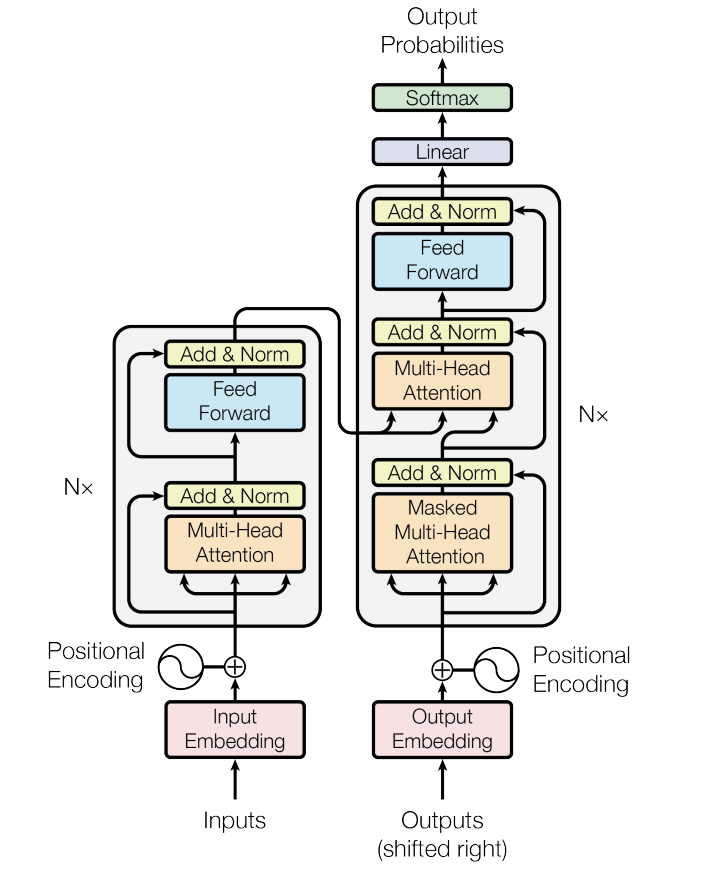}
    \caption{ The Transformer - model architecture}
    \label{fig:transformer}
\end{figure}

Building on the power of attention, \citep{Vaswani2017Attention} proposed the Transformer architecture (Figure \ref{fig:transformer}) in the landmark paper “Attention Is All You Need.” By stacking multi-head self-attention layers interleaved with position-wise feedforward networks, the Transformer dispensed with recurrence entirely. Each token in the input attends to every other token in parallel, enabling both efficient GPU utilization and effective modeling of long-range dependencies. Positional encodings were introduced to inject sequence order information, ensuring that the model remained sensitive to word order despite its parallel structure.

The GPT series adapted the Transformer into a decoder-only format optimized for generative pre-training. GPT-1 demonstrated that a unidirectional Transformer, pre-trained on a large corpus and then fine-tuned on task-specific data, could achieve strong performance across multiple benchmarks \citep{Radford2018GPT}. GPT-2 scaled this idea by increasing model size and training data, yielding emergent few-shot learning abilities without task-specific fine-tuning \citep{Radford2019GPT2}. GPT-3 further amplified scale—using 175 billion parameters—and showed that in-context learning alone could lead to high-quality performance on tasks ranging from question answering to code generation \citep{Brown2020GPT3}.

In each of these stages, improvements in model architecture, training regimes, and scale have collectively enhanced the ability of LLMs to generate coherent, contextually appropriate text. These foundational innovations set the stage for the subsequent emergence of agentic systems, where LLMs are no longer confined to single-turn generation but instead engage in multi-step planning, tool use, and interactive behaviors that reflect increasingly complex, autonomous capabilities.

\section{The Transformer Revolution and the GPT Lineage}

Since the debut of the original GPT decoder‐only model, a series of architectural and training refinements have steadily improved both the efficiency and capabilities of large language models. We organize these innovations into five “steps” of evolution:

\begin{enumerate}[label=\textbf{Step \arabic*:}, leftmargin=*, align=left]
  \item \textbf{Foundation – GPT Decoder Architecture.}  
    The GPT series \citep{Radford2018GPT,Radford2019GPT2,Brown2020GPT3} introduced the notion that a unidirectional (“causal”) Transformer decoder, pre‐trained on massive text corpora via next‐token prediction, could serve as a general-purpose language engine. Unlike the encoder–decoder setup of the original Transformer, the GPT decoder stacks masked self‐attention layers followed by position‐wise feed-forward networks, enabling autoregressive generation without the need for explicit encoder outputs. This simple yet powerful design demonstrated that, with sufficient scale, language models could perform downstream tasks through zero- and few-shot prompting alone.

  \item \textbf{Alignment via Reinforcement Learning.}  
    With GPT’s generative prowess came concerns over unaligned or unsafe outputs. Reinforcement Learning from Human Feedback (RLHF) \citep{Ouyang2022InstructGPT} addressed this by treating the language model as a policy in a reinforcement learning framework. Human raters first score model outputs for desired qualities; these scores train a reward model, which in turn guides policy updates via Proximal Policy Optimization. RLHF elevated the “human-likeness” and safety of LLM responses, becoming standard in instruct-tuned variants like InstructGPT and ChatGPT.

  \item \textbf{Accuracy Optimizations.}  
    As models scaled, small architectural tweaks yielded outsized gains in stability and performance:
    \begin{itemize}
      \item \textbf{RMSNorm.} Unlike LayerNorm, which relies on both mean and variance, Root Mean Square normalization (RMSNorm) scales activations solely by their root mean square. This reduces computational overhead and mitigates exploding gradients when training on extremely large datasets or with large batch sizes.\citep{Xiong2023PreRMSNorm}
      \item \textbf{SwiGLU Activation.} Replacing the standard GeLU or ReLU in feed-forward blocks, the SwiGLU function gates half the hidden units using a sigmoid, enabling more expressive, conditional computation in the MLP sub-layers without significantly increasing parameter count.\citep{Shazeer2020GLU}
      \item \textbf{RoPE (Rotary Positional Encoding).} Traditional absolute positional encodings can struggle with generalizing to longer sequences. Rotary Positional Encoding injects relative positional information via rotation matrices applied to query/key projections, yielding better extrapolation to unseen sequence lengths and improved accuracy on long-context tasks.\citep{Su2021RoFormer}
    \end{itemize}

  \item \textbf{Training Efficiency.}  
    To reduce compute without sacrificing quality:
    \begin{itemize}
      \item \textbf{Grouped-Query Attention (GQA).} GQA clusters attention heads into small groups (e.g., eight heads per group), sharing key and value projections within each group. This design cuts computation by 6–8× with only a minor (10–20\%) impact on perplexity, making larger models more practical to deploy.\citep{Ainslie2023GQA}
      \item \textbf{Ghost Attention.} For conversational agents, maintaining a persistent “system” prompt or persona over long dialogues is critical. Ghost Attention passes a lightweight version of the system token through each block as a “ghost,” ensuring core instructions remain salient even after thousands of tokens of user and assistant history.\citep{Touvron2023Llama2}
    \end{itemize}

  \item \textbf{Scaling and Advanced Optimizations.}  
    As context lengths and model sizes grew, new techniques preserved performance and efficiency:
    \begin{itemize}
      \item \textbf{Ultra-Long Context Windows.} Methods like sliding-window attention, chunked processing, and sparse global attention enable models to process tens or hundreds of thousands of tokens in a single pass, critical for document-level understanding and multi-document reasoning.\citep{Beltagy2020Longformer}
      \item \textbf{Document-Aware Masking.} During pre-training, isolating attention to within-document token groups prevents the model from conflating patterns across separate documents, improving coherence on multi-text corpora.\citep{Saha2021PositionMasking}
      \item \textbf{Document-Level Pre-Norm + SwiGLU Decoder.} Applying layer normalization before attention and feed-forward sub-layers (pre-norm) on document chunks, combined with SwiGLU activations in the decoder, stabilizes gradients over longer contexts and accelerates convergence on large-scale corpora.\citep{Xiong2020LayerNorm}
      \item \textbf{Mixture of Experts (MoE).} By routing each token through only a small subset of “expert” feed-forward modules, MoE layers achieve model capacity equivalent to trillions of parameters with the compute cost of a much smaller dense model, yielding dramatic reductions in FLOPs and memory use.\citep{Shazeer2017MoE}
      \item \textbf{iRoPE (Improved RoPE).} Extending Rotary Positional Encoding to handle extremely long contexts, iRoPE modifies the rotational frequency schedule to maintain stable relative positional representations even when context windows exceed tens of thousands of tokens.\citep{Liu2025VRoPE}
    \end{itemize}
\end{enumerate}

Together, these steps allows basic transformer to become today’s highly efficient, and scalable foundation for agentic LLM systems

\section{Instruction Tuning \& the Birth of AI Assistants}

InstructGPT marked OpenAI’s first large‐scale foray into creating a human‐like assistant from GPT-3, transforming a general‐purpose text generator into a conversational policy that reliably follows user instructions. By layering human feedback on top of GPT-3’s generative capabilities, InstructGPT became the inaugural system on OpenAI’s API explicitly designed to behave as a helpful, safe, and aligned assistant. The development unfolded in three key stages:

\begin{enumerate}[label=\textbf{\arabic*.}, leftmargin=*, align=left]
  \item \textbf{Supervised Fine‐Tuning with Human Demonstrations.} Expert annotators authored thousands of prompt–completion pairs (“demonstrations”) illustrating the desired assistant behavior (e.g., concise answers, polite tone). GPT-3 was fine‐tuned on this dataset via maximum‐likelihood training, resulting in an initial policy that prioritized instruction adherence over the freeform continuation of text.

  \item \textbf{Preference Collection and Reward Modeling.} Annotators then ranked multiple model outputs for the same prompt according to helpfulness, correctness, and safety. These pairwise preferences were used to train a reward model that predicts human judgments, serving as a learned objective for further optimization.

  \item \textbf{Reinforcement Learning from Human Feedback (RLHF).} Treating the fine‐tuned policy as an initial agent, OpenAI applied Proximal Policy Optimization (PPO) to maximize the reward model’s scores while penalizing divergence from the supervised baseline. This RL phase refined the policy to generate responses more closely aligned with human preferences, reducing harmful or irrelevant content.
\end{enumerate}

Human evaluations showed that, despite having 100× fewer parameters than vanilla GPT-3 (1.3 B vs.\ 175 B), InstructGPT outputs were consistently preferred by raters—demonstrating that instruction tuning via human feedback can yield a more useful and safe assistant than simply scaling model size (Ouyang et al., 2022; OpenAI, 2022). This milestone marks the true beginning of the personal assistant paradigm in large language models.

\section{Chain-of-Thought Reasoning: Concepts and Milestones}

Before examining fully agentic systems, it is instructive to trace the development of chain‐of‐thought (CoT) prompting, the seminal technique that first unlocked multi‐step reasoning in large language models. Whereas standard prompting elicits a direct answer, CoT intersperses intermediate natural‐language reasoning steps between the problem statement and the final response. A typical few‐shot CoT prompt consists of triples of the form $\langle\text{input},\;\text{chain\_of\_thought},\;\text{output}\rangle$, where the “chain\_of\_thought” is a human‐crafted sequence of logical steps leading from the input to the answer.

CoT prompting exhibits several key properties:
\begin{itemize}
  \item \textbf{Decomposition of Complex Tasks.} By breaking multi‐step problems (e.g., arithmetic word problems or multi‐hop inference) into explicit sub‐steps, CoT allows the model to allocate additional computation where long reasoning chains are required.
  \item \textbf{Interpretability and Debuggability.} The natural‐language intermediate steps provide a transparent “window” into the model’s reasoning process, enabling practitioners to inspect and correct logical missteps or hallucinations.
  \item \textbf{Task Generality.} Any problem solvable by humans through language—commonsense reasoning, symbolic manipulation, logical deduction—can, in principle, be tackled via CoT prompting.
  \item \textbf{Ease of Elicitation.} With sufficiently large models (on the order of 100 B parameters or more), simply including a few exemplars of reasoning chains in a prompt reliably elicits multi‐step reasoning without additional fine‐tuning.
\end{itemize}

However, CoT prompting only yields performance gains in very large models. Models with fewer than $\sim$10 B parameters generate fluent but logically unsound chains, often harming overall accuracy. Moreover, CoT is sensitive to prompt wording: small changes in exemplar phrasing can significantly impact performance, necessitating careful prompt engineering.

To address these challenges, several refinements have emerged:
\begin{itemize}
  \item \textbf{Zero‐Shot CoT.} Prompting with a generic trigger (e.g., “Let’s think step by step.”) induces reasonable reasoning traces in large‐scale models without any examples.
  \item \textbf{Auto‐CoT.} Automates exemplar selection by clustering unlabeled questions using Sentence‐BERT embeddings and k‐means; representative questions from each cluster serve as prompts, with their chains generated via Zero‐Shot CoT to form a refined few‐shot prompt.
  \item \textbf{Self‐Consistency.} Samples multiple reasoning paths for the same problem and takes a majority vote over final answers, improving robustness at the cost of increased compute proportional to the number of samples.
  \item \textbf{Tree of Thoughts.} Explores a tree of partial reasoning states by branching at decision points, evaluating multiple continuations in parallel, and pruning low‐quality paths, effectively performing a search over reasoning trajectories.
\end{itemize}

Together, these innovations transformed CoT from a few‐shot curiosity into a robust, automated framework for eliciting and verifying complex reasoning in LLMs, laying the groundwork for fully autonomous agentic architectures.

\section{Instilling CoT Mechanisms in Foundation Models}

Large‐scale LLMs do not naturally produce coherent, step‐by‐step reasoning; instead, they learn to predict the next token, which can lead to plausible but ungrounded “fluent” outputs. To imbue them with reliable internal reasoning traces—i.e., chain‐of‐thought (CoT) behavior—models like DeepSeek V2 undergo a three‐stage training regimen specifically designed to bake in their own “thinking.”\citep{Liu2024DeepSeekV2}

\begin{enumerate}[label=\textbf{Stage \arabic*:}, leftmargin=*, align=left]
  \item \textbf{Foundational Pre‐training.}  
    The model is first exposed to an enormous and diverse corpus (DeepSeek V2: 8.1 trillion tokens), with a deliberate emphasis on text that exhibits logical structure—mathematical proofs, code repositories, and formally written arguments. By learning to predict masked or next tokens in these high‐logic domains, the model internalizes patterns of structured reasoning (e.g., inference chains, algorithmic steps) at scale. This phase does not explicitly teach reasoning, but builds the raw capacity to represent and manipulate abstract symbols and sequences, providing the substrate for later CoT instruction.

  \item \textbf{Supervised Fine‐Tuning for Reasoning (SFT‐CoT).}  
    To convert latent reasoning abilities into explicit, verbalized CoT traces, the model is fine‐tuned on a curated dataset of problem–solution pairs annotated with detailed intermediate steps. Each example employs markup (e.g., \texttt{<think>…</think>}) around the reasoning segment, guiding the model to generate its internal monologue when solving a task. For instance, a math word problem is paired not merely with an answer but with a fully fleshed‐out sequence of natural‐language sub‐calculations. Through maximum‐likelihood training on this data, DeepSeek V2 learns that high‐quality responses must “show their work,” effectively learning the mapping:
    \[
      \text{Prompt} \;\longrightarrow\; \langle\!\!\texttt{<think>…(intermediate steps)…</think>}\!\!\rangle \;\text{Final Answer}.
    \]
    This supervised stage is critical: without explicit reasoning demonstrations, even very large models will default to direct answers or produce illogical chains.

  \item \textbf{Reinforcement Learning Polishing.}  
    Having learned to output reasoning traces, the model’s next objective is to refine their quality and alignment with human judgments. DeepSeek V2 employs Group Relative Policy Optimization (GRPO), a variant of PPO tailored for group‐based evaluation. During each RL episode:
    \begin{enumerate}[label=\arabic*., leftmargin=*, align=left]
      \item The model generates \(k\) candidate answers, each paired with its CoT sequence.
      \item A human‐trained reward model evaluates each full candidate (reasoning + answer) on dimensions such as logical coherence, correctness, and clarity.
      \item GRPO updates the policy to increase the probability of the highest‐scoring candidates, while constraining deviation from the behavior learned during SFT.
    \end{enumerate}
    This RL phase does not introduce new reasoning patterns; rather, it sharpens the model’s preference toward its most human‐preferred chains, reducing spurious or verbose detours and aligning its reasoning style with annotated examples.
\end{enumerate}

Together, these stages transform a generic next‐token predictor into an LLM that internally reasons in natural language, providing transparent, debug‐friendly chains of thought that underpin more advanced agentic behaviors (e.g., planning, tool selection, self‐critique)

\section{Emergence of Agentic Architectures}

Large language models have evolved far beyond one-shot text generators into interactive agents that can reason, act, and even learn from their own mistakes. Three pioneering frameworks—ReAct, Toolformer, and Reflexion—illustrate this transformation by weaving together natural-language “thoughts,” actions via external tools, and iterative self-improvement.

\begin{description}[leftmargin=*, style=nextline]
  \item[\textbf{ReAct.}] ReAct first demonstrated that an LLM could alternate between reasoning and tool use within a single prompt. Instead of separating planning from execution, ReAct interleaves a \emph{Thought} (the model’s chain-of-thought), an \emph{Action} (a tool or API call), and an \emph{Observation} (the tool’s response). For example:
  \begin{verbatim}
Thought: The Eiffel Tower is in Paris.
Action: wiki_api("Paris country")
Observation: France
Thought: The capital of France is Paris.
Answer: Paris
  \end{verbatim}
  By giving the model explicit permission to think and act in tandem, ReAct achieves marked improvements on multi-hop question answering, interactive simulations, and task-oriented benchmarks—while also producing a transparent trace that facilitates debugging and interpretability.\citep{Yao2022ReAct}

  \item[\textbf{Toolformer.}] Building on the idea that models can teach themselves to call tools, Toolformer takes a self-supervised approach. It begins with just a few example API calls, then lets the model generate raw text and hypothesize where additional calls might help. These proposed calls are executed, and only those that reduce the model’s perplexity are retained. Finally, the model is fine-tuned on this enhanced corpus—now enriched with in-line tool calls and their results. The result is an LLM that learns, almost autonomously, when and how to use calculators, search engines, or translators, delivering zero-shot performance gains across QA, arithmetic, and translation tasks.\citep{Schick2023Toolformer}

  \item[\textbf{Reflexion.}] While ReAct and Toolformer focus on mixing reasoning with action, Reflexion adds a third dimension: iterative self-critique. Reflexion agents operate in cycles of Actor, Evaluator, and Reflector. The Actor attempts the task, generating its reasoning and actions; the Evaluator then judges success or failure; if the attempt fails, the Reflector produces a natural-language reflection—pointing out errors and suggesting a better approach. This reflection is appended to the prompt, and the Actor tries again. By repeating this loop and storing past reflections in memory, Reflexion agents progressively refine their behavior, achieving dramatic gains on benchmarks like complex QA and code generation—all without any parameter updates.\citep{Shinn2023Reflexion}
\end{description}

We also want to highly main methodology which is used among agents which is RAG. Retrieval-Augmented Generation (RAG) is a framework that enhances the performance of large language models (LLMs) by integrating external knowledge retrieval into the generation process. Unlike standard LLMs, which rely solely on their internal parameters to produce responses, RAG dynamically retrieves relevant documents or facts from an external knowledge base—such as a vector database, search index, or API—before generating the final output. This hybrid approach addresses key limitations of LLMs, including hallucinations, outdated knowledge, and insufficient domain-specific expertise. By grounding responses in retrieved information, RAG improves factual accuracy, contextual relevance, and adaptability to specialized tasks. In practice, RAG is commonly used for tasks like question answering, code generation, and summarization, where up-to-date or domain-specific information is critical. It also reduces the need for extensive fine-tuning, making LLM-powered systems more efficient and cost-effective while enabling continuous knowledge updates without retraining the entire model.

Together, these three frameworks chart a clear trajectory: from simple prompt-driven reasoning, through self-supervised tool mastery, to fully autonomous, self-improving agents. They form the foundation of today’s agentic LLM systems—capable not only of generating coherent text but of planning, executing, and learning within dynamic environments.

\section{Modular Orchestration: Multi-Module \& Autonomous Agents}

\begin{figure}[h]
    \centering
    \includegraphics[width=0.8\textwidth]{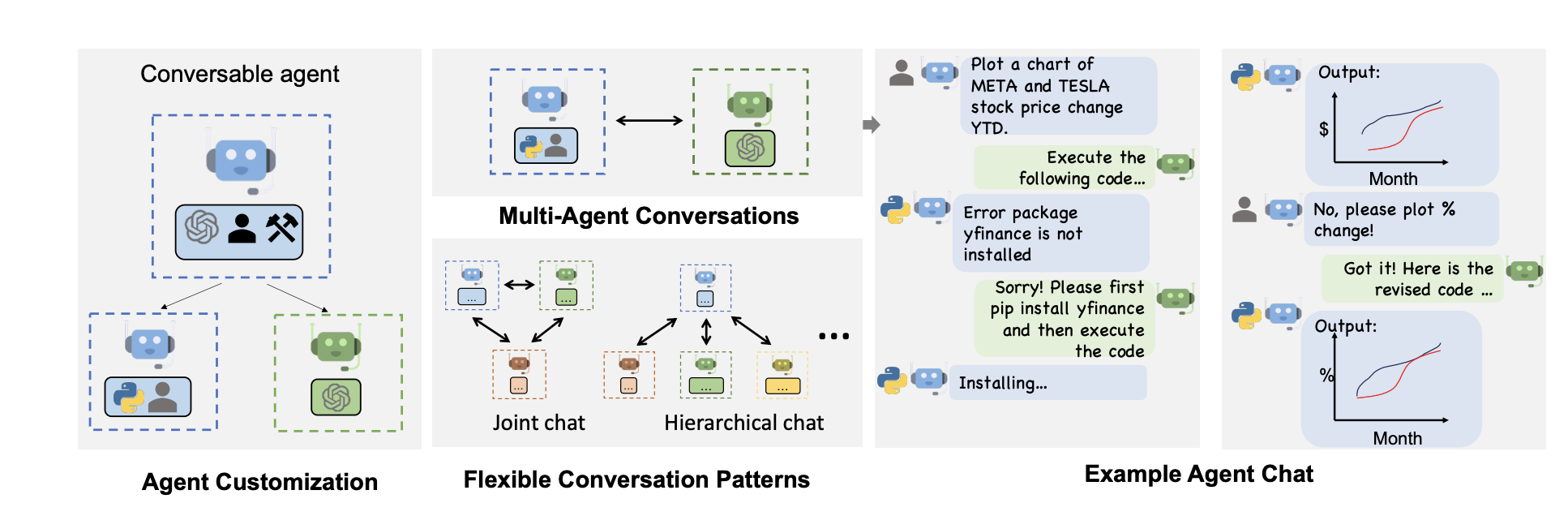}
    \caption{AutoGen enables diverse LLM-based applications using multi-agent conversations. (Left) AutoGen agents are conversable, customizable, and can be based on LLMs, tools, humans, or even a combination of them. (Top-middle) Agents can converse to solve tasks. (Right) They can form a chat, potentially with humans in the loop. (Bottom-middle) The framework support flexible
conversation patterns.}
    \label{fig:autogpt}
\end{figure}

Beyond the paradigms of ReAct, Toolformer, and Reflexion, recent systems demonstrate how LLMs can function as central controllers, coordinating a diverse ecosystem of expert modules or even managing entire workflows end‐to‐end. Three prominent examples are HuggingGPT, Gorilla, and AutoGPT—each illustrating a different point along the spectrum of agentic complexity.

\begin{description}[leftmargin=*, style=nextline]
  \item[\textbf{AutoGPT.}]  
    AutoGPT (\ref{fig:autogpt} Figure) represents one of the first widely adopted autonomous agents built around LLMs. AutoGPT employs GPT‐4 or GPT‐3.5 to autonomously pursue user‐defined goals through a continuous loop of planning, execution, observation, and reflection \citep{richards2023autogpt}. Starting with a high‐level objective (e.g., “draft a marketing plan”), AutoGPT generates a to‐do list of subtasks, executes each via model calls or external tools (e.g., web search, file I/O), captures the outputs as observations stored in a vector‐embedding memory, and iteratively refines its plan based on those observations. This self‐driving workflow requires no additional human prompts once initiated, demonstrating that LLMs can operate as project managers, orchestrating complex, multi‐step processes with minimal oversight.
  \item[\textbf{HuggingGPT.}]  
    HuggingGPT leverages a large language model (e.g., ChatGPT) as a task planner and router for hundreds of specialized AI models in the Hugging Face ecosystem. When given a user request, HuggingGPT first decomposes it into subtasks expressed in natural language, then selects appropriate expert models (vision, speech, text, etc.) based on their metadata, executes each model to obtain results, and finally aggregates those outputs into a coherent response. This approach demonstrates that an LLM can effectively manage a heterogeneous, multi‐modal AI pipeline, achieving strong performance across complex, cross‐domain benchmarks and paving a path toward modular AGI \citep{li2023hugginggpt}.

  \item[\textbf{Gorilla.}]  
    Gorilla extends this concept by training the LLM itself to generate structured, JSON‐style function calls for a vast API surface—covering cloud services, databases, and domain‐specific tools. During training, Gorilla ingests API schemas and usage examples, learning to predict not only when a tool should be invoked but also how to format its arguments correctly. At inference, the model emits API call requests directly, which are then executed and re‐injected into the prompt as observations. Gorilla’s ability to natively support thousands of functions with minimal prompt engineering significantly outperforms even GPT‐4 on API‐calling accuracy in question answering and code generation tasks \citep{li2023gorilla}.
\end{description}

Together, these architectures chart an evolution from single‐agent reasoning and tool use to multi‐module orchestration and fully autonomous task execution. They underscore a growing consensus: the next generation of robust AI agents will combine the generative and planning capabilities of LLMs with specialized expert modules—seamlessly coordinated under a unified “language‐based” control plane.

\section{Evaluating Agentic LLMs: Frameworks \& Metrics}

Assessing the performance of large language models—and, in particular, their agentic extensions—requires a multi‐tiered strategy. Traditional NLP metrics provide a baseline for static text generation, but dynamic, tool‐augmented, and multi‐turn behaviors necessitate specialized benchmarks and agent‐centric evaluation frameworks.

\begin{enumerate}[label=\textbf{\arabic*.}, leftmargin=*, align=left]
  \item \textbf{Classical Evaluation Metrics.}  
    Early model assessment borrowed directly from supervised NLP and machine learning. Standard techniques include k‐fold cross‐validation and held‐out test sets to measure generalization. For generation tasks, string‐oriented metrics such as BLEU \citep{Papineni2002BLEU} and ROUGE \citep{Lin2004ROUGE} quantify n‐gram overlap with reference outputs, while token‐level accuracy and F1 score gauge precise prediction in classification or sequence‐tagging tasks. Although these metrics remain valuable for gauging fluency and surface‐level correctness, they exhibit fundamental limitations:
    \begin{itemize}
      \item \textbf{Determinism Assumption:} BLEU and ROUGE reward exact overlaps, yet many valid generations diverge lexically, penalizing creative or paraphrastic output.
      \item \textbf{Lack of Interaction Modeling:} Token‐level scores assume a one‐to‐one mapping between input and output and cannot capture iterative refinement or decision‐making processes central to agentic behavior.
      \item \textbf{No Tool or Memory Assessment:} These metrics ignore the agent’s ability to call external tools, maintain state across turns, or leverage retrieval.
    \end{itemize}

  \item \textbf{Static LLM Benchmark Suites.}  
    To probe general‐purpose LLM capabilities, large‐scale collections of tasks have been assembled:
    \begin{itemize}
      \item \textbf{MMLU} evaluates performance across 57 academic and professional subjects in few‐shot settings \citep{Hendrycks2021MMLU}.
      \item \textbf{HELM} measures model behavior across toxicity, robustness, and fairness dimensions via standardized prompts \citep{Li2022HELM}.
      \item \textbf{TruthfulQA} targets knowledge‐based reliability by exposing hallucination through questions with known ground truth \citep{Lin2022TruthfulQA}.
      \item \textbf{BIG-bench} assembles hundreds of probing tasks—from symbolic manipulation to cultural reasoning—to stress‐test emerging abilities \citep{Srivastava2022BIGBench}.
    \end{itemize}
    While these suites reveal latent LLM capacities and scaling trends, they share critical shortcomings:
    \begin{itemize}
      \item \textbf{Single‐Shot Evaluation:} Each prompt yields exactly one static answer, whereas agentic systems plan, revise, and interact with environments over multiple steps.
      \item \textbf{No External Tool Integration:} Benchmarks assume purely text‐based reasoning and do not evaluate API invocation, database queries, or code execution.
      \item \textbf{Homogeneous Knowledge Scope:} Tasks cover broad domains and cannot validate performance on narrowly defined, proprietary, or dynamically evolving knowledge bases.
    \end{itemize}

  \item \textbf{Why These Benchmarks Fail for Agentic, Niche Applications.}  
    Agentic systems embody autonomy: they perceive, decide, and act—often under uncertainty—across extended interactions. Current benchmarks remain tethered to static text completion:
    \begin{itemize}
      \item \textbf{No Feedback Loops:} Agentic evaluation must simulate environments (e.g., web search, database updates, user feedback) and measure belief or plan updates.
      \item \textbf{Absence of Goal‐Directed Tasks:} Benchmarks rarely specify long‐horizon objectives requiring subtask decomposition, resource management, or conditional execution.
      \item \textbf{Inability to Test Domain‐Specific Expertise:} Deploying agents in specialized fields demands benchmarks built on precise corpora and task structures.
    \end{itemize}
\end{enumerate}

\section{Related Work}

The rapidly expanding field of LLM-based agent evaluation has garnered significant attention. Notably, agentic systems based on LLMs and their diagnostic methodologies represent a novel and frontier research area. The majority of papers in this domain are currently in pre-print state and date between 2024 and 2025. Our research indicates that the majority of leading laboratories currently engaged in this problem, leading to the emergence of both comprehensive surveys and highly specialized benchmarks. For instance, \citep{Yehudai2025SurveyLLMAgents} present the first systematic survey of evaluation methodologies for agentic systems. They categorize existing work along four dimensions: core agent capabilities, application-specific benchmarks, generalist agent assessments, and evaluation frameworks. They advocate for the development of finer-grained, scalable, and safety-aware evaluation methods.

\subsection*{Tool Use and Function-Calling Evaluation}

Evaluating an agent’s ability to invoke and coordinate external tools has spurred a suite of benchmarks. \citep{Qin2023APIBank} curate API-Bank, a large corpus of real-world API calls interleaved with natural text, to measure API detection, sequencing, and parameter accuracy. \citep{Song2023RestBench}  release RestBench, focusing on multi-API workflows driven by user instructions, and demonstrate that agents frequently misuse endpoints or fail to sequence calls correctly. More recently, \citep{Zhong2025ComplexFuncBench} present ComplexFuncBench, designed to assess nested API interactions under constraint-heavy scenarios, revealing that existing models struggle with implicit parameter inference and chained dependencies.

\subsection*{Self-Reflection and Memory Assessment}

Beyond planning and tool use, agents must monitor their own reasoning and store context over extended dialogues. \citep{Li2024ReflectionBench} propose Reflection-Bench, which probes cognitive reflection by evaluating an agent’s ability to detect and adjust for reasoning errors, counterfactuals, and meta-reflection steps. In parallel, \citep{Huet2025EpisodicMemoryBenchmark} develop an episodic memory benchmark that uses synthetic narratives to measure how accurately agents generate, retrieve, and update event memories during multi-turn interactions.

\subsection*{Unit Tests for Agent Evaluation}

Inspired by software engineering practices, recent work has begun to propose unit-test-style evaluation for LLM agents. \citep{AgentUnit} introduce AgentUnit, a framework that decomposes complex tasks into discrete functional units and evaluates each via predefined test cases, demonstrating improved diagnostic precision in failure analysis. Similarly, \citep{Test4LLM} propose Test4LLM, which automatically generates unit tests from task specifications and measures correctness, robustness, and edge-case handling; they show that incorporating unit test feedback during training leads to significant gains in reliability across diverse benchmarks.

Despite these advances, existing evaluations often isolate individual capabilities rather than measuring their interplay in fully autonomous, multi-agent workflows.

\subsection*{Agent-as-a-Judge and Instruction-Following Frameworks}

\citep{Zhuge2024AgentAsJudge} extend the notion of “LLM-as-a-Judge” to Agent-as-a-Judge, wherein one agentic system evaluates another by providing step-wise feedback and scoring on complex multi-step tasks; they show that this meta-evaluation outperforms static LLM-based judges and approaches human reliability on code generation tasks. More recently, \citep{Qi2025AgentIF} introduce AgentIF, the first benchmark targeting instruction-following in realistic agentic scenarios; AgentIF compiles long, constraint-laden prompts from industrial and open-source agents and reveals that even advanced LLMs fail to adhere to critical tool and conditional constraints.\\

Despite the advancements in evaluation methodologies, a significant gap remains in addressing the “why” and “how to improve” aspects of system performance. Currently, most evaluations rely on three primary techniques: creating a benchmark for specific tasks, conducting unit tests of inputs to agents and their expected outputs, and utilizing human or LLM as judges for task evaluations. While these techniques effectively track when systems are incorrect, they fail to provide insights into the underlying reasons and potential areas for improvement, which are essential for the diagnostic process.
Diagnostics hold greater significance for expert systems compared to general problem solvers like ChatGPT or Google Gemini. Examples of specialist systems include legal advisors, lawyers, doctors who prescribe medications, financial advisors, and neurodiversity and gender language specialists, all of which are discussed in this paper. Our framework and methodology are specifically designed to address diagnostic cases of this nature by guiding system towards expert like behavior and diagnostic reasons of difference between real expert and system.

\chapter{Methodology}
\noindent This chapter formalizes the diagnostic method introduced conceptually in Chapter~2 into an implementable pipeline. Building on the limitations of static, single-pass benchmarks discussed in \S~2.8--2.9, we define two orthogonal diagnostics---Extraction (ED) and Behaviour (BD)---and the data/agent infrastructure required to run them reproducibly. The outputs here seed the use case in Chapter~4 and provide the scoring, prescription, and mapping machinery referenced by the results in Chapter~5.

\section{Agent Diagnostics}

Agentic LLMs emerge primarily to overcome two intertwined challenges: the limited capacity of a single context window and the need for specialized, domain-robust behavior. As context windows grow to encompass hundreds or thousands of tokens, disparate conversational threads—say, dog breeds, combustion chemistry, astrophysics, and bicycling techniques—become interleaved. Much like how a Generative Adversarial Network (GAN) trained on mixed datasets might produce images that blend incongruous features (e.g., a dog with fire textures, a bicycle orbiting a black hole), an LLM confronted with too many unrelated topics will “mutate” its internal representation in unpredictable ways, manifesting as semantic drift or hallucinations.

To illustrate, consider prompting a single LLM session with successive requests about HR policy, astrophysics, and recipe recommendations. As each new topic is appended, residual attention weights from earlier segments can bleed into later responses, causing the model to conflate concepts—mixing jargon, inventing spurious links, or contradicting itself. In practice, this leads to an elevated hallucination rate and reduced reliability once the context window approaches its capacity (Ji et al., 2023).

To mitigate context-induced failure modes, practitioners have partitioned responsibility across domain-specific agents. For example, an “HR Agent” handles hiring policy, a “Physics Agent” answers astrophysical queries, and a “Culinary Agent” generates recipes. Each agent’s context window remains focused on a narrow ontology, avoiding cross‐domain interference. This modularization parallels ensemble methods in machine learning: specialized experts operate in isolation, and a gating mechanism routes queries to the appropriate module.

Paradoxically, the same mechanism that causes unwanted drift can be harnessed to steer agent behavior. Imagine constructing a “spectrum” of ten hypothetical HR specialists, ranging from novice to expert. By seeding the prompt with exemplar dialogues from different points along this spectrum, one can interpolate or extrapolate the LLM’s language to emulate an intermediate proficiency. In effect, the context becomes a “behavioral latent space” wherein small perturbations nudge the agent’s decisions, tone, and knowledge depth toward a target expert profile

\section{Agent Diagnostic Method for Expert Systems (ADM--ES)}
\label{sec:method}

\begin{figure}[h]
    \centering
    \includegraphics[width=0.4\textwidth]{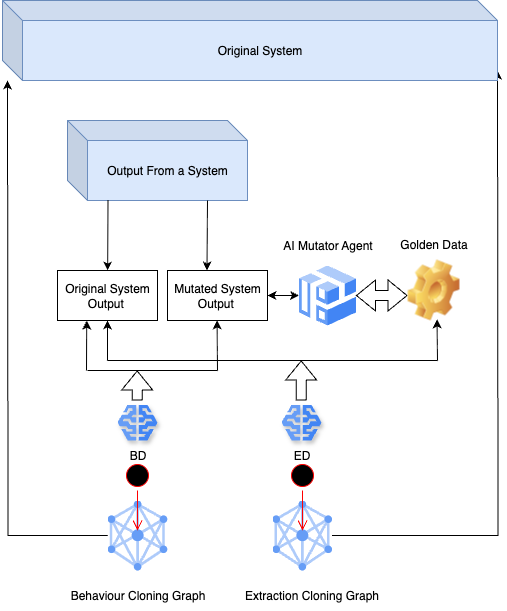}
    \caption{Text Block Extraction Diagnostic (ED)
and Behaviour Diagnostic (BD) Diagram.}
    \label{fig:diagnostic_diagram_method}
\end{figure}

This section presents a reproducible diagnostic method for steering and evaluating agentic expert systems along two orthogonal axes which you can see on \ref{fig:diagnostic_diagram_method} Diagram: (i) \textbf{Text Block Extraction Diagnostic (ED)} and (ii) \textbf{Behaviour Diagnostic (BD)}. The goal is to direct an LLM-based system under test (SUT) toward a desired \emph{extraction style}, \emph{tone}, and \emph{answering behaviour} by (1) curating a compact \emph{golden dataset} of expert demonstrations, (2) generating a broad \emph{mutated silver dataset} via an \emph{Agent Mutator} that RAG-conditions on golden exemplars and performs CoT-guided behaviour transfer, and (3) deploying an \emph{ED Agent} and \emph{ED Agent} that scores and proposes prompt/system improvements.

We denote: input document/context $x$, extracted sentences $E(x)$, expert rationale/thoughts $T(x)$, final answer $y$, and environment context $C$. The golden dataset is $\mathcal{G} = \{(x_i, E^*_i, T^*_i, y^*_i)\}_{i=1}^n$, where $(\cdot)^*$ indicates expert annotation.

\subsection{Overview of Pipeline}
\label{sec:pipeline}

\noindent\textbf{Stage A --- Golden curation.} Experts annotate small, high-fidelity triples $(E^*, T^*, y^*)$ per task instance $x$.\\

\noindent\textbf{Stage B --- Silver mutation.} Given a new $x$ and $C$, retrieve top-$k$ behaviourally similar golden exemplars with a vector index; condition a mutation LLM with CoT instructions to transform $(x,\ \text{SUT output})$ into expert-like $(\tilde{T}, \tilde{y})$ while preserving task semantics; admit examples by BERTScore-based quality checks; aggregate into $\mathcal{S}$ (silver).\\

\noindent\textbf{Stage C --- Agent Judge.} A rubric-prompted LLM evaluates: (i) extraction quality by comparing $E_{\text{SUT}}(x)$ vs.\ $E^*(x)$, and (ii) behaviour (tone, reasoning clarity, adherence to expert style) by comparing $(T_{\text{SUT}}, y_{\text{SUT}})$ vs.\ silver $(\tilde{T}, \tilde{y})$. The Judge also emits concrete \emph{prompt/system deltas} to incline the SUT toward target behaviour.\\

\noindent\textbf{Stage D --- Prescription Map} This stage are focus in creating of a map of prescriptions for behavior and extraction diagnostic and creation of clusters for all main groups of prescriptions.\\

\subsection{Text Block Extraction Diagnostic (ED)}
\label{sec:ed}

In this subsection, we will provide an explanation of the concept of Extraction Diagnostic and define all its main components. Please note that Extraction Diagnostic is conducted between the golden dataset and the target system’s response.

\subsubsection{Task formalization}
\label{sec:ed-task}

Let $E(x) = \{e_j\}$ denote the set of extracted sentences from $x$. We evaluate \emph{sentence selection only}.

\begin{itemize}
  \item \textbf{Sentence selection.} Ground truth $E^*(x)\subseteq S(x)$ where $S(x)$ is the sentence set. Metrics: sentence-level precision/recall/F1
  \item \textbf{Salience ranking (optional).} If experts provide ranked importance, compute NDCG@$p$ (Normalized Discounted Cumulative Gain) between SUT and expert orders.
\end{itemize}

\subsubsection{Golden construction protocol}
\label{sec:ed-gold}

\begin{enumerate}
  \item \textbf{Sampling.} Stratify by domain, length, and difficulty to avoid bias; target $n\in[100, 500]$ high-quality items.
  \item \textbf{Annotation.} For each $x$:
  \begin{itemize}
    \item $E^*$: minimal sufficient set of sentences to support $y^*$.
    \item $T^*$: terse, stepwise rationale explaining \emph{why} each $e\in E^*$ is selected (chain-of-thought style but concise).
    \item $y^*$: the canonical expert response (style-controlled).
  \end{itemize}
  \item \textbf{Quality control (optional).} Dual annotation with adjudication; compute inter-annotator agreement coefficients (e.g., Krippendorff’s $\alpha$, Cohen’s $\kappa$); disagreements can be resolved by a lead annotator.
  \item \textbf{Schema.} Store JSONL records: \texttt{\{id, domain, x, E*, T*, y*, policy\_tags\}}.
\end{enumerate}

\subsubsection{ED metrics}
\label{sec:ed-metrics}

\begin{itemize}
  \item \textbf{Extraction:} sentence-level precision/recall/F1
  \item \textbf{Rationale alignment (Optional):} similarity between $T_{\text{SUT}}$ and $T^*$ using BERTScore-F1
\end{itemize}

\subsection{Agent Mutator: Silver Dataset Generation }
Agent Mutator serves as the primary component responsible for transferring behavior between experts and systems. Its main objective is to significantly expand the dataset without the involvement of experts, thereby accelerating the process.  

\label{sec:mutator}

\subsubsection{Inputs and retrieval}
\label{sec:mutator-inputs}

Let $f$ be the embedding model and $\mathbb{V}$ a vector DB (e.g., Qdrant). For a query representation $q=f(x, C, \text{SUT}(x))$, retrieve $R_k=\{(x_r,E^*_r,T^*_r,y^*_r)\}_{r=1}^k$ by cosine similarity; optionally hybridize with BM25 for lexical grounding. Typical $k\in[3,8]$.

\subsubsection{Mutation objective}
\label{sec:mutator-objective}

Produce $(\tilde{T}, \tilde{y})$ such that:
\begin{enumerate}
  \item \textbf{Behavioural alignment:} $(\tilde{T}, \tilde{y})$ match style/tone and reasoning patterns of $R_k$.
  \item \textbf{Semantic preservation:} $\tilde{y}$ addresses $x$ under $C$ without leaking $x_r$ specifics.
  \item \textbf{Non-copying:} limit verbatim overlap with any $y^*_r$ (e.g., \textsc{SelfBLEU} $<\tau_{\text{copy}}$).
\end{enumerate}

\subsubsection{Mutation prompt (sketch)}
\label{sec:mutator-prompt}

\begin{quote}\itshape
You are mutating the SUT’s output to match expert behaviour. Given (a) the task input and environment context, (b) the SUT’s original thoughts/output, and (c) $k$ nearest expert exemplars (with their rationales), \textbf{rewrite} the thoughts and the final answer to emulate the experts’ reasoning style, tone, and structure. Preserve all task-specific facts from the input; do not copy exemplar phrasings.
\end{quote}

\subsubsection{Quality check \& acceptance (mean BERTScore over \texorpdfstring{$k$}{k})}
\label{sec:mutator-qc}

We compute BERTScore as the \emph{average} over all $k$ retrieved exemplars rather than the maximum.

\paragraph{Behaviour mutation check.}
\begin{equation}
\overline{\mathrm{BERTScore}}_y = \frac{1}{k}\sum_{r=1}^k \operatorname{BERTScoreF1}(\tilde{y}, y^*_r),\quad
\overline{\mathrm{BERTScore}}_T = \frac{1}{k}\sum_{r=1}^k \operatorname{BERTScoreF1}(\tilde{T}, T^*_r).
\end{equation}
Accept a mutated instance if both averages fall within $[\tau_{\min}, \tau_{\max})$ with $\tau_{\max}<1$ to avoid cloning; maintain diversity via pairwise embedding distance $\overline{d}>\delta$.

\paragraph{Extraction check.} For ED we compute \emph{sentence-level BERTScore} by comparing the SUT’s extracted sentences $E_{\text{SUT}}(x)$ to expert $E^*(x)$ (aligned at sentence granularity) and report the \emph{average} score. This complements the discrete metrics in \S\ref{sec:ed-metrics}.

\subsubsection{Output schema}
\label{sec:mutator-schema}

Silver record $(x, \tilde{T}, \tilde{y}, R_k, \text{qc\_metrics})$ with provenance and thresholds; store in $\mathcal{S}$.

\subsection{Behavior Diagnostic (BD): Scoring \& Steering}
\label{sec:judge}

This subsection will elucidate the concept of behavior diagnostic between an expert and a system that you are targeting. 

\subsubsection{Judge roles}
\label{sec:judge-roles}

\begin{enumerate}
  \item \textbf{ED scoring.} Compare $E_{\text{SUT}}$ vs.\ $E^*$ with sentence-level metrics; summarize as $\mathrm{EDScore}\in[0,100]$.
  \item \textbf{BD scoring.} Compare $(T_{\text{SUT}}, y_{\text{SUT}})$ vs.\ $(\tilde{T}, \tilde{y})$ and rubric-score tone, structure, factuality, and reasoning clarity.
  \item \textbf{Steering suggestions.} Emit structured \emph{deltas}: \texttt{\{system\_prompt\_edits, tool\_policies, decoding\_params, retrieval\_params, guardrails\}}.
\end{enumerate}

\subsubsection{Rubric for BD (0--5 per facet; weighted sum)}
\label{sec:judge-rubric}
Weights for rubrics can specifically be tuned to author preference to achieve more precise results.
\begin{itemize}
  \item \textbf{Factual sufficiency ($w{=}0.3$):} Does the answer rest entirely on selected evidence?
  \item \textbf{Reasoning clarity ($w{=}0.25$):} Are steps minimal, correct, and traceable?
  \item \textbf{Tone/style match ($w{=}0.2$):} Formality, hedging discipline, persona adherence.
  \item \textbf{Extraction--answer coherence ($w{=}0.15$):} Each claim is supported by $E$.
  \item \textbf{Tool-use fidelity ($w{=}0.1$):} Correct tool selection and parameterization if applicable.
\end{itemize}

Calibrate the Judge via pairwise comparisons against human ratings on $\mathcal{G}$, optimizing prompt variants to maximize Kendall’s $\tau$ with experts 

\subsubsection{Judge outputs}
\label{sec:judge-outputs}

\begin{itemize}
  \item \textbf{Scores:} \texttt{EDScore}, \texttt{BDScore}, confidence.
  \item \textbf{Rationales:} concise justifications (stored, not exposed to end-users at inference).
  \item \textbf{Prescriptions:} actionable edits (e.g., add extraction policy bullets; adjust temperature; tighten retrieval $k$; include verifiability line).
\end{itemize}

\subsection{Metrics Summary}
\label{sec:metrics-summary}

By summarizing everything which was explained before we should get this picture.

\noindent\textbf{Extraction (ED).} Sentence-level precision/recall/F1, Jaccard@sentence, ROUGE-Lsum; NDCG for ranked salience; NLI sufficiency.

\noindent\textbf{Behaviour (BD).}
\begin{itemize}
  \item \textbf{Similarity:} BERTScore-F1 between SUT and silver; compute the \emph{average} across all $k$ retrieved exemplars (no max pooling).
  \item \textbf{Style fit:} classifier-based tone/formality score vs.\ target band; LM perplexity under expert-style language model.
  \item \textbf{Reasoning quality:} length-normalized CoT edit distance to silver; error taxonomies from Judge.
\end{itemize}

\noindent\textbf{Mutation quality.} mean-BERTScore acceptance window; \textsc{SelfBLEU} for copying; coverage/diversity across domains.

\noindent\textbf{Reliability.} Bootstrap 95\% CIs over items; report run-to-run variance under fixed seeds.

\subsection{Recommendation Map (UMAP) of Prescriptions}
\label{subsec:recommendation-map}

After each diagnostic round, the \textit{Agent Judge} outputs a prescription (structured edits, failure tags, and context). We embed these prescriptions and construct a similarity graph to obtain a low-dimensional map (2D/3D) of recurring improvement themes. This map aggregates many fine-grained recommendations into a few major \emph{blocks} that generalize across examples and supports interactive inspection and auditing.

\subsubsection{Record schema.}
Each recommendation $r_i$ is persisted as
\[
\texttt{\{suggestion\_id, source\_item\_id, prescription, failure\_tags, embedding\}}
\]
with embedding $\mathbf{z}_i \in \mathbb{R}^d$ computed over the serialized recommendation (\textit{prescription\_delta + failure\_tags + context\_signature}). 

\subsubsection{Pipeline.}
\begin{enumerate}
  \item \textbf{Persist recommendation records.} For each judged case, append the record $\texttt{r}_i$ and compute $\mathbf{z}_i$.
  \item \textbf{Build a $k$-NN similarity graph.} Let $V=\{1,\dots,n\}$ index recommendations and
  \[
    \mathcal{N}_k(i) = \operatorname*{arg\,top\text{-}k}_{j\neq i}\ \cos(\mathbf{z}_i,\mathbf{z}_j).
  \]
  Create edges $E=\{(i,j): j\in \mathcal{N}_k(i)\}$ with base weights $w^{(0)}_{ij}=\cos(\mathbf{z}_i,\mathbf{z}_j)$.
  \item \textbf{UMAP projection.} Apply UMAP to the weighted graph / embeddings $\{\mathbf{z}_i\}$ to obtain $\mathbf{y}_i \in \mathbb{R}^p$ with $p\in\{2,3\}$:
  \[
    \Phi:\ \mathbb{R}^d \to \mathbb{R}^p,\qquad \mathbf{y}_i=\Phi(\mathbf{z}_i).
  \]
  \item \textbf{Cluster to form major blocks.} Cluster $\{\mathbf{y}_i\}$ with HDBSCAN/DBSCAN or $k$-means to obtain labels $\ell_i \in \{1,\dots,B\}$ (major recommendation blocks).
\end{enumerate}

\subsubsection{Improvements tracking}

We also maintain historical evaluation deltas per system improvement recommendation. We assume that as system is improved, we will see a lower and lower number of recommendations:
\[
\Delta ED_i = ED^{\text{before}}_i - ED^{\text{after}}_i,\qquad
\Delta BD_i = BD^{\text{before}}_i - BD^{\text{after}}_i,
\]
and a normalized improvement score
\[
\mathrm{Imp}(i) \;=\; \alpha\,\widetilde{\Delta ED}_i \;+\; \beta\,\widetilde{\Delta BD}_i,\quad \alpha,\beta \ge 0,\ \alpha+\beta=1,
\]
where tildes denote normalization across the dataset.

The UMAP map reduces the search space for improvements by consolidating many micro-prescriptions into a handful of stable, reusable blocks that generalize across training and evaluation examples. The optional improvement-aware reweighting promotes recommendations with proven $\Delta ED/\Delta BD$ gains


\subsection{Implementation Notes}
\label{sec:impl-notes}

\begin{itemize}
  \item \textbf{Stack.} FastAPI service; Qdrant for corpus/storage and vectors; Pydantic for data validation for RAG and Judge pipelines.
  \item \textbf{Schemas.} Versioned JSONL with \texttt{source\_ids},
  \item \textbf{Safety.} Red-team prompts against overfitting to silver style; constrain CoT exposure in deployment (use rationale-lite in production).
\end{itemize}

\subsection{Limitations \& Threats to Validity}
\label{sec:limitations}

\begin{itemize}
  \item \textbf{Judge bias \& over-alignment.} Mitigate via calibration on expert gold and pairwise comparisons; diversify Judge models.
  \item \textbf{Silver drift.} Periodically refresh $\mathcal{S}$ with new gold; monitor acceptance window and diversity metrics.
  \item \textbf{Domain shift.} Validate across domains and difficulty buckets; include OOD tests.
\end{itemize}

By closing the loop—mutating context toward desired behavior, diagnosing the system’s response, and adapting the mutation strategy—we aim to operationalize a reproducible, domain‐agnostic diagnostic protocol. This protocol not only surfaces latent cognitive failures (e.g., context degradation, tool misrouting) but also prescribes concrete interventions for elevating an LLM agent’s competency to an expert level.

\noindent We specified ADM--ES as a four-stage loop: curate high-fidelity gold (\S~3.2.2), generate expert-aligned silver via controlled mutation (\S~3.2.3), and judge/steer agents with a rubric-based critic that emits actionable prescriptions (\S~3.2.4). We summarized ED/BD metrics (\S~3.2.5) and described how prescriptions are embedded into a reusable recommendation map (\S~3.2.6). These mechanisms define the test harness that Chapter~4 instantiates over JobFair's recruiter-assistant agents and that Chapter~5 uses to quantify behavioural transfer and extraction fidelity.

\chapter{Diagnostic of Agentic System}
This chapter applies ADM--ES (Chapter~3) to the JobFair system. We state exactly which agents are evaluated with which diagnostics, and what is not covered. The production pipeline contains six agents; this study focuses on the two bias-mitigation specialists that directly influence inclusive phrasing and clarity:

\begin{itemize}
  \item Gendered Language Agent (GLA)
  \item Neurodiversity Agent (NDA)
\end{itemize}

\noindent \textbf{Diagnostics coverage:}
\begin{itemize}
  \item \textbf{BD (Behaviour Diagnostic)} --- implemented for GLA and NDA via the Mutator + BD Agent loop over large job-description corpora; we mutate only the \emph{recommendation} and \emph{example} fields while keeping the \texttt{bias\_item} fixed, then score stylistic/tonal alignment and prescribe steering edits.
  \item \textbf{ED (Extraction Diagnostic)} --- implemented for GLA and NDA on the golden set to test sentence-level extraction fidelity against expert annotations.%
  \item \textbf{Recommendation Map} --- built from the ED and BD prescriptions emitted during BD/ED runs; used later for clustering recurrent failure modes and steering changes across releases.
\end{itemize}

\noindent \textbf{Implemented vs.\ general methodology.} From Chapter~3, we implemented:
\begin{itemize}
  \item \textbf{Stage A (gold curation)} across the two bias categories, producing expert \texttt{bias\_item} annotations with rationales and exemplar rewrites.
  \item \textbf{Stage B (silver mutation)} only for BD and only on \emph{recommendation}/\emph{example} fields; acceptance thresholds were documented but not enforced in this iteration to observe raw capability.
  \item \textbf{Stage C (Agent Judge)} for BD (GLA, NDA) and ED (NDA), producing facet scores and structured prescriptions. The ED Judge operated exclusively on the expert-annotated golden set.
  \item \textbf{Stage D (Recommendation Map)} --- we cluster the recommendation map for behaviour steering.
\end{itemize}

\noindent We instantiated ADM--ES on two specialist agents within JobFair. BD was executed for GLA and NDA by mutating recommendations/examples and scoring stylistic alignment; ED was executed for GLA and NDA on the expert-annotated golden set. The outputs of these runs (scores and prescriptions) feed the recommendation map and set up the quantitative analyses reported in Chapter~5.

\section{JobFair Agentic System Architecture}

\begin{figure}[h]
    \centering
    \includegraphics[width=0.8\textwidth]{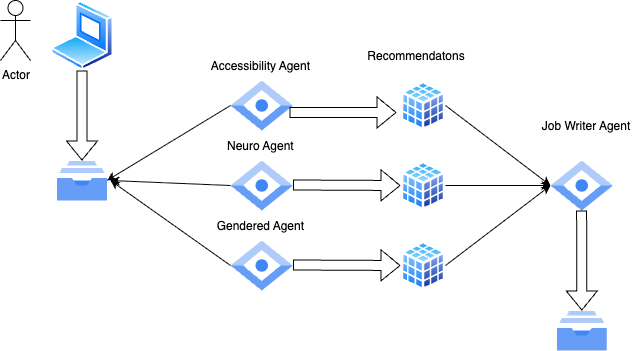}
    \caption{Simplified Version of JobFair System}
    \label{fig:job_fair_sys}
\end{figure}

The JobFair system (Figure \ref{fig:job_fair_sys}) is designed as a modular, multi‐agent architecture whose primary aim is to analyze and optimize job description templates for inclusive language. By detecting and correcting bias along gender, neuro‐diversity, and age dimensions, the system helps organizations craft postings that attract a diverse pool of qualified candidates and accurately represent the role and company culture.

At runtime, a raw job description template or draft is routed through a pipeline of six specialized agents. Each agent operates on a shared context store but maintains a domain‐specific model and prompt strategy. After sequential processing, a final “Job Writer” agent synthesizes all revisions into a polished posting.

\subsubsection*{1. Job Description Agent}
The Job Description Agent generates the initial draft by applying the JobFair's proprietary logic—drawing on historical posting performance data, standard competency frameworks, and role‐specific templates. It sets the baseline content and structure for downstream refinement.

\subsubsection*{2. Accessibility Agent}
The Accessibility Agent specializes in enhancing readability and usability for candidates with disabilities (e.g., visual impairments, dyslexia). Leveraging guidelines and plain‐language best practices, it:
\begin{itemize}
  \item Simplifies complex sentences and jargon  
  \item Ensures sufficient contrast and alt‐text placeholders for embedded media  
  \item Normalizes formatting (e.g., consistent heading hierarchy, bullet lists)  
\end{itemize}
By injecting accessibility‐focused annotations and rewriting instructions, this agent increases compliance and broadens the candidate pool.

\subsubsection*{3. Gendered Language Agent (Focus)}
The Gendered Language Agent detects and neutralizes gender‐coded words and phrases. Using a lexicon of gender‐biased terms (e.g., “aggressive,” “nurturing”) and contextual embedding techniques, it:
\begin{itemize}
  \item Flags strongly gender‐polymorphous descriptors  
  \item Suggests neutral or balanced alternatives (e.g., “assertive” instead of “aggressive”)  
  \item Estimates a “gender bias score” to quantify imbalances  
\end{itemize}
This targeted intervention aligns with best practices in inclusive recruitment and reduces the risk of inadvertently discouraging applicants of any gender.

\subsubsection*{4. Neuro‐diversity Agent (Focus)}
The Neuro‐diversity Agent ensures the language and structure accommodate neuro‐divergent applicants (e.g., those with autism spectrum conditions, ADHD). Drawing on clinical‐informed recommendations, it:
\begin{itemize}
  \item Minimizes figurative language and implicit expectations  
  \item Encourages explicit role expectations, clear instructions, and transparent evaluation criteria  
  \item Introduces “checklist” sections to reduce cognitive overload  
\end{itemize}
By embedding these features, the agent fosters clarity and predictability, which have been shown to improve application rates and candidate experience among neuro‐divergent populations.

\subsubsection*{5. Requirement Analysis Agent}
The Requirement Analysis Agent parses the draft to extract and validate core competencies, qualifications, and experience levels. It cross‐references standardized occupational taxonomies to ensure that listed requirements are neither overly restrictive nor too vague.

\subsubsection*{6. Job Writer Agent}
Finally, the Job Writer Agent consolidates all revisions into a coherent, publication‐ready job posting. It harmonizes tone, removes redundant edits, and applies final stylistic consistency checks before output.

\vspace{1ex}
\normalsize\textbf{Component Focus:} In this work, we concentrate on 3 (Gendered Language), and 4 (Neuro‐diversity), as these directly influence and create building blocks for Job Writer Agent to create final draft of the job description

\section{Golden Data Set and Pre‐processing}

\subsection{Expert Judgments Data - Golden Dataset}

Company‐recruited PhD‐level specialists at the intersection of Computer Science, Data Science, and Cognitive Psychology to create a curated “golden” data set of sentence‐level annotations. Experts reviewed a curated corpus of 13 real‐world job descriptions with $+-30$ sentences for each job description manually extracted every sentence deemed relevant to one of three agent roles: (1) Accessibility Agent, (2) Gendered Language Agent, and (3) Neuro‐diversity Agent.

\subsubsection*{Creation of the Golden Data Set}  
Each expert operated on an Excel template with the following columns:
\begin{itemize}
  \item \texttt{sentence\_text}: full UTF-8 text of the extracted sentence.
  \item \texttt{comments, expert thoughts}: free‐text notes or rationale.
  \item \texttt{agent\_category}: one of \{\texttt{accessibility}, \texttt{gendered\_language}, \texttt{neurodiversity}\}.
  \item \texttt{Replacement Example}: example of preferred job description.
\end{itemize}

After completion, the Excel workbook (13 sheets, one per job description) was exported to CSV and parsed via a custom Python script. The script performed:
\begin{enumerate}
  \item \textbf{Field normalization}: ensured consistent column names, trimmed whitespace, and validated UTF-8 encoding.
  \item \textbf{Identifier generation}: concatenated \texttt{job\_desc\_id} and \texttt{sentence\_id} to form a globally unique \texttt{annotation\_id}.
  \item \textbf{JSON serialization}: emitted one JSON object per row, using the schema below.
\end{enumerate}

\subsubsection*{JSON Schema for Expert Annotations}  
\begin{verbatim}
{
    "bias_item": "full UTF-8 text of the extracted sentence",
    "Expert_Comments": "free‐text notes or rationale.",
    "recommendation": "example of preferred job description.",
    "expert_id": "expert_001 - id of expert",
    "job_id": "job_010 - id of job description"
},
\end{verbatim}

In total, experts annotated 156 sentences across the 13 descriptions: 33 for accessibility, 31 for gendered language, and 92 for neuro‐diversity. This JSON corpus constitutes our “golden” reference for downstream model evaluation and prompt‐engineering experiments.

\subsection{Job Description Data}

To evaluate our agentic system on real‐world inputs, we collected a corpus of 300 publicly available job descriptions from Indeed via a custom web scraper. These descriptions cover a diverse range of industries and seniority levels, ensuring broad coverage for downstream analysis.

\subsubsection*{Data Ingestion Stage}  
\begin{itemize}
  \item A Python‐based scraper (built on \texttt{requests} and \texttt{BeautifulSoup}) navigates Indeed job listing pages, paginating through search results by keyword and location.  
  \item For each listing, the scraper captures the raw HTML payload—including all embedded CSS and JavaScript—and logs metadata (URL, timestamp, job\_id) into a Supabase table \texttt{raw\_job\_descriptions}.  
  \item Duplicate detection is performed via URL fingerprinting to ensure each of the 300 entries is unique.
\end{itemize}

\subsubsection*{Pre‐processing Stage}  
\begin{itemize}
  \item \textbf{HTML/CSS/JS Stripping}: Each \texttt{raw\_job\_description.html} field is parsed with \texttt{BeautifulSoup} to remove all \textless script\textgreater{} and \textless style\textgreater{} blocks, then all HTML tags are stripped to yield plain text.  
  \item \textbf{Boilerplate Removal}: Common headers, footers, and navigation elements are identified via regular expressions and removed to isolate the actual job description content.  
  \item \textbf{Normalization}: The remaining text is whitespace‐normalized (collapsing multiple spaces and line breaks), Unicode‐normalized to NFC, and trimmed.  
  \item \textbf{Storage}: Cleaned plain‐text descriptions are written back into JSONL and Supabase in a table \texttt{cleaned\_job\_descriptions} with fields:
    \begin{itemize}
      \item \texttt{job\_id} (primary key),
      \item \texttt{clean\_text} (UTF-8 string),
      \item \texttt{word\_count}, 
      \item \texttt{ingestion\_date}.
    \end{itemize}
\end{itemize}

After these two stages, the 300 cleaned job descriptions are ready for sentence‐level segmentation and subsequent analysis of performance by our diagnostic systems, as well as for automated processing by the JobFair agentic pipeline. 

\subsubsection*{JSON Schema for Job Description}  
\begin{verbatim}
{
    "job_id": "id of the job inside of vector DB",
    "job_description": "plain text of job-description",
    "vector_embedding": "vector embedding for job over job_description",
},
\end{verbatim}

\subsection{Vector Embedding Stage For Golden Dataset and Job Description}  
\begin{itemize}
  \item \textbf{Model Selection}: The \texttt{sentence-transformers/all-MiniLM-L6-v2} \citep{all-minilm-l6-v2} model (384-dimensional dense embeddings) was used due to its balance between computational efficiency and semantic accuracy for short to medium-sized texts such as job descriptions. The model was loaded using the \texttt{SentenceTransformers} Python library.
  
  \item \textbf{Encoding Full Job Descriptions and bias item}: For each \texttt{job\_descriptions} field in the job\_descriptions JSONL file and \texttt{bias\_item}, the entire entity was fed into the transformer model as a single sequence. The output embedding is a fixed-size vector $\mathbf{v} \in \mathbb{R}^{384}$ representing the overall semantic meaning of the job description.
    \begin{itemize}
      \item Input text is tokenized, truncated to a maximum of 256 tokens (model constraint), and processed through the MiniLM encoder.
      \item The [CLS] token representation is pooled and normalized via L2 normalization to ensure uniform vector scales.
    \end{itemize}

  \item \textbf{Batch Processing}: Embeddings were generated in batches of size 32 for efficient inference. Computation was executed on CPU with \texttt{float32} precision to balance speed and memory usage.

  \item \textbf{Storage \& Schema}: The resulting embeddings are stored as serialized \texttt{float32} arrays in two locations:
    \begin{enumerate}
      \item Supabase, in the \texttt{job\_embeddings} and \texttt{golden dataset}  table,
      \item A vector database (Qdrant) for semantic search and clustering.
    \end{enumerate}
\end{itemize}

\section{Mutation for BD Diagnostic}
\label{sec:stage43-bd-mutation}

At the outset of system development, the two original agents in the JobFair pipeline---\textbf{Gendered Language}, \textbf{Neurodiversity}, and \textbf{Accessibility}---produced outputs consisting of only two fields: \texttt{bias\_item} and \texttt{recommendation}. As the multi-agent pipeline evolved, later agents introduced an additional \texttt{example} field to provide a concrete rewritten segment of the job description. For confidentiality reasons, we cannot disclose the full agentic pipeline beyond this description. For consistency in this study, we extracted the outputs of the two target agents and expanded them into a unified schema of \texttt{bias\_item}, \texttt{recommendation}, and \texttt{example} before applying the mutation process.

\subsection{Resulting JSON structure.} For each processed job description:
\begin{verbatim}
{
  "bias_item": "sentence extracted from job description",
  "recommendation": "system’s recommended replacement or intervention",
  "example": "system-generated example of a rewritten job description segment"
}
\end{verbatim}

To ensure a consistent evaluation pipeline, only the \textbf{recommendation} and \textbf{example} fields were targeted for mutation in this stage. The original \texttt{bias\_item} served as the contextual anchor and was never modified.

\subsection{Inputs and Retrieval}
The mutation pipeline began by embedding each system output record using the \textsc{MiniLM} encoder. For every candidate record, the embedding of its \texttt{bias\_item} was used to retrieve the top-$k$ most similar expert annotations from the golden dataset stored in \textsc{Qdrant}. These exemplars---containing expert rationales and preferred replacements---served as behavioural guides for the mutation model.

\subsection{Mutation Objective}
The goal of this stage was to transform JobFair system suggestions into \emph{silver recommendations} that better align with expert style and tone while preserving semantic fidelity. Specifically:
\begin{itemize}
  \item \textbf{Behavioural Alignment:} Mutated recommendations and examples should emulate expert phrasing and intervention strategies.
  \item \textbf{Semantic Preservation:} All domain-specific content and contextual meaning from the \texttt{bias\_item} must remain intact.
  \item \textbf{Controlled Diversity:} Outputs must be distinct from both the original system suggestion and the retrieved exemplars, reducing risks of overfitting or memorization.
\end{itemize}

\subsection{Mutation Process}
A chain-of-thought prompted LLM (mutation model \ref{ref:agent_mutator}) was conditioned on:
\begin{enumerate}
  \item the raw job description;
  \item the system's original \texttt{recommendation} and \texttt{example};
  \item $k$ retrieved golden exemplars with expert rationales.
\end{enumerate}
The model was instructed to rewrite only the \texttt{recommendation} and \texttt{example} fields, generating outputs that mirror expert reasoning and intervention style. The prompt explicitly constrained the model from altering the \texttt{bias\_item} and discouraged verbatim copying.

\subsection{Quality Check \& Acceptance}
Mutated outputs were validated against expert exemplars using \emph{mean BERTScore} (applied separately to \texttt{recommendation} and \texttt{example}). We haven't applied acceptance criteria as we wanted to test raw capability but potential future work can test were:
\begin{itemize}
  \item Scores must fall within a calibrated similarity window $[\tau_{\min}, \tau_{\max})$.
\end{itemize}

\subsection{Output Schema}
Each silver record was stored in JSON format with provenance and metrics:
\begin{verbatim}
{
  "bias_item": "unaltered input sentence",
  "job_id": "id of job description",
  "mutated_recommendation": "rewritten expert-like recommendation",
  "mutated_example": "expert-style replacement example",
  "quality_metrics": {
    "bert_score_recommendation": 0.xx,
    "bert_score_example": 0.xx,
  }
}
\end{verbatim}

\subsection{Contribution to BD Diagnostic}
The silver records produced in this stage constitute the \emph{behavioural diagnostic (BD) silver dataset}. By mutating only the \texttt{recommendation} and \texttt{example} fields, the process directly targets the \emph{core prescriptive components} of the JobFair Agent's outputs. This enables the Agent Judge to:
\begin{itemize}
  \item evaluate the alignment of recommendations with expert judgement;
  \item detect failure modes in phrasing, tone, or inclusivity that the raw system outputs overlooked;
  \item provide more reliable signals for iterative system improvement across accessibility, gendered language, and neurodiversity diagnostics by extrapolating golden dataset to bigger scale.
\end{itemize}

\noindent In summary, this stage operationalizes behaviour-focused mutation over system suggestions, yielding a scalable silver dataset that anchors BD diagnostics in expert-informed yet systematically diversified outputs.

\section{Agent Judge for BD Diagnostic}

Next stage of the diagnostic pipeline applies the \textbf{Agent Judge} to the silver dataset generated in Stage~C. While the mutation process produced expert-like recommendations and examples, the Judge operationalizes their evaluation, quantifying the degree of alignment between the JobFair system outputs and the expert-informed silver standard.\\

\textbf{Inputs.} Each silver record contained (i) the unaltered \texttt{bias\_item}, (ii) the JobFair system’s original \texttt{recommendation} and \texttt{example}, and (iii) the mutated expert-style \texttt{mutated\_recommendation} and \texttt{mutated\_example}. These were embedded and passed to the Judge for scoring.

\textbf{Judge Structure.} The Agent Judge was divided into two complementary parts to ensure both contextual fidelity and targeted evaluation:
\begin{itemize}
    \item \textbf{System Prompt (:\ref{ref:agent_bd_prompt}) } For each diagnostic run, the Judge was primed with the original system prompt of the corresponding specialist agent (Neurodiversity, Gendered Language, or Accessibility). This preserved the intended operational persona of the evaluated agent, aligning the Judge’s reasoning with the same domain-specific framing.
    \item \textbf{Action Prompt.} The action prompt contained: (a) the original job description text, (b) the original system-generated \texttt{recommendation} and \texttt{example}, and (c) the corresponding silver-standard \texttt{mutated\_recommendation} and \texttt{mutated\_example}. This structure provided the Judge with both the raw system output and its expert-informed counterpart for direct comparison.
\end{itemize}

\textbf{Evaluation Rubric.} The Judge applied a weighted rubric with five facets:

\begin{itemize}
  \item \textbf{Tone Match} (Weight: 0.25): replicates emotional, rhetorical, and formality tone. \emph{Scale 0–5; anchors: 0 inverted/inappropriate; 1–2 major mismatch; 3 broadly aligned; 4 well-matched; 5 indistinguishable.}
  \item \textbf{Stylistic Fidelity} (Weight: 0.25): similarity in syntax, vocabulary register, and rhythm. \emph{Scale 0–5; anchors: 0 entirely different; 1–2 significant simplification/mismatch; 3 generally similar; 4 close match; 5 near-exact imitation.}
  \item \textbf{Manner of Expression} (Weight: 0.25): preserves rhetorical/expressive style (abstraction, metaphor, indirectness). \emph{Scale 0–5; anchors: 0 opposite strategy; 1–2 major shift; 3 partially retained; 4 good retention; 5 full mirroring.}
  \item \textbf{Semantic Alignment} (Weight: 0.25): preserves meaning, logical structure, and intent. \emph{Scale 0–5; anchors: 0 reversed/major distortion; 1–2 partial with omissions/additions; 3 core retained; 4 relations intact; 5 complete fidelity.}
\end{itemize}

The weights were evenly distributed to put equal emphasis on all aspects of behavior diagnosis.

Each facet was rated on a 0--5 scale, aggregated into a \textbf{BDScore} for behaviour diagnostics.

\textbf{Outputs.} For each case, the Judge produced:
\begin{enumerate}
    \item \textbf{Scores:} numerical ED and BD scores, with BD focusing on recommendation--example fidelity.
    \item \textbf{Rationales:} concise justifications for high or low facet scores.
    \item \textbf{Prescriptions:} structured recommendations for system steering (e.g., tighten retrieval $k$, increase formal tone discipline, reduce hedging in recommendations).
\end{enumerate}

\textbf{Contribution.} By contrasting JobFair system outputs with silver-standard mutations, the Agent Judge exposed failure modes invisible to static evaluations, including:
\begin{itemize}
    \item recommendations that preserved semantics but failed stylistically (e.g., overly informal phrasing);
    \item examples that drifted from inclusivity guidelines despite lexical similarity;
    \item inconsistencies between bias\_item and replacement suggestions.
\end{itemize}

Crucially, the Judge’s prescriptions were stored as vectorized recommendations in the \textbf{recommendation map}, enabling clustering of recurrent failures into major improvement blocks. This allowed the evaluation process not only to measure performance but also to steer iterative refinement of the JobFair system toward expert-level behaviour.

\section{Agent Judge for ED Diagnostic}

In contrast to the behaviour-focused BD diagnostic, the \textbf{ED (Extraction Diagnostic)} evaluation employed the Agent Judge over the curated golden dataset. Here, the aim was to assess whether the JobFair system correctly identified and extracted the same sentences that domain experts had annotated as critical for bias detection.\\

\textbf{Inputs.} The ED Judge operated exclusively on the original set of 13 job descriptions that were annotated by experts to create the golden dataset. The JobFair system-generated extractions for the 13 golden job descriptions, they were compared directly against these expert annotations. Each golden record contained the \texttt{bias\_item} field, serving as the canonical ground truth for sentence-level extraction.\\

\textbf{Judge Structure.} As with the BD diagnostic, the ED Judge was divided into two complementary prompts:\\
\begin{itemize}
    \item \textbf{System Prompt (\ref{ref:agent_bd_prompt})} The Judge was initialized with the original system prompt of the agent under evaluation (Neurodiversity, Gendered Language, or Accessibility), ensuring that the Judge operated under the same contextual framing as the JobFair extraction agents.
    \item \textbf{Action Prompt.} The action prompt contained: (a) the full job description text, (b) the JobFair system’s extracted \texttt{bias\_item(s)}, and (c) the expert-annotated golden \texttt{bias\_item(s)}. This setup enabled a direct, sentence-level comparison between system output and expert ground truth.\\
\end{itemize}

\textbf{Evaluation Rubric.} As evident, the evaluation rubrics presented here differ from those in the methodology section. They have been intentionally modified to assess the system’s capability to operate effectively without sentence similarity tools, as outlined in the methodology section, future study can make unhancement in this area. For ED diagnostics, the Judge rubric is focused on extraction fidelity:
\begin{itemize}
    \item \textbf{Correctness} (Weight: 0.35): accuracy of extracted findings vs.\ gold; labels and polarity/negation must match.
    \item \textbf{Completeness} (Weight: 0.30): recall of expert findings; missing diagnoses or clinical findings reduce the score.
    \item \textbf{Over-Extraction} (Weight: 0.15): penalizes extra or hallucinated findings not present in gold.
    \item \textbf{Detail Accuracy} (Weight: 0.10): precision of attributes (severity, stage, onset/timing, test values, locations, etc.).
    \item \textbf{Terminology Consistency} (Weight: 0.05): uses expert-preferred terms or exact synonyms; ambiguous/wrong terms penalized.
    \item \textbf{Reasoning Alignment} (Weight: 0.05): cited evidence/support aligns with expert-cited evidence; mismatched rationale gets partial credit.
\end{itemize}

Weights were distributed to put greater power on Correctness and Completeness which JobFair team was mostly interested in.

Each dimension was rated and aggregated into an \textbf{EDScore} (0–100), representing sentence-level extraction accuracy.\\

\textbf{Outputs.} For each job description, the ED Judge produced:
\begin{enumerate}
    \item \textbf{Scores:} Correctness, Completeness, Over-Extraction and a normalized EDScore.
    \item \textbf{Rationales:} brief explanations for missing or spurious extractions.
    \item \textbf{Prescriptions:} targeted adjustments to extraction policies (e.g., “expand coverage to capture boundary sentences,” “reduce false positives on stylistic phrases”).
\end{enumerate}

\textbf{Contribution.} By grounding the evaluation in the golden dataset, the ED Judge provided a robust benchmark for extraction fidelity. Unlike the BD stage, which assessed style and tone alignment via silver-standard mutations, the ED stage strictly measured whether the JobFair system could recover the same key evidence sentences that experts had identified. This enabled detection of:
\begin{itemize}
    \item missed extractions (false negatives) that omitted crucial bias indicators,
    \item over-extraction (false positives) where irrelevant sentences were included,
    \item inconsistencies in how the system parsed structurally complex job descriptions.
\end{itemize}

Together, these evaluations allowed the ED Judge to serve as a foundational diagnostic

\section{Recommendation Map }
\label{sec:recommendation-map}

The final stage of the diagnostic pipeline aggregated Agent Judge prescriptions into a structured, navigable space. Each prescription---defined as a combination of suggested edits, identified failure tags, and contextual metadata---was embedded into a dense vector representation. This embedding enabled the construction of a similarity graph over all accumulated recommendations across the JobFair system.

\subsection{Record schema.}
Each prescription was persisted in JSON format as a structured record:
\begin{verbatim}
{
  "suggestion_id": "unique identifier",
  "source_item_id": "link to original system item",
  "prescription": "structured recommendation text",
  "failure_tags": ["list of diagnostic failure categories"],
  "embedding": "vector encoding of the prescription"
}
\end{verbatim}

\noindent The embedding was derived by serializing the prescription text and associated failure tags into a unified representation, then computing a dense vector encoding via the same \textsc{MiniLM} model used in earlier retrieval stages.

\subsection{Pipeline.}
\begin{enumerate}
  \item \textbf{Persist records.} Every diagnostic case generated a prescription that was serialized and stored alongside its embedding.
  \item \textbf{Construct graph.} A $k$-nearest-neighbour graph was built over cosine similarities of embeddings, producing a weighted adjacency structure that connected semantically related recommendations.
  \item \textbf{UMAP projection.} The high-dimensional space of recommendations was projected into two or three dimensions using UMAP, which preserved local neighbourhoods while enabling visualization and interactive analysis.
  \item \textbf{Cluster formation.} The projected embeddings were clustered using HDBSCAN or $k$-means, yielding coherent blocks of prescriptions that corresponded to recurring failure patterns or intervention strategies.
\end{enumerate}

The resulting recommendation map reduced the complexity of analysis by consolidating numerous fine-grained prescriptions into a smaller number of major thematic clusters. This visual and structural aggregation enabled systematic auditing of system weaknesses, revealing patterns of stylistic failure, recurrent inclusivity issues, or domain-specific misalignments. Beyond offering a diagnostic snapshot, the map served as a reusable knowledge structure: future system evaluations could be contextualized against these clusters, ensuring consistency and comparability across multiple rounds of diagnostics.

We instantiated BD and ED for GLA and NDA, producing facet scores and structured prescriptions under the unified bias\_item/bias\_category schema. The outputs here (scores + prescriptions) populate the recommendation map and are analysed quantitatively in Chapter 5.

\chapter{Results}
This chapter reports quantitative outcomes from the Chapter 4 runs. We first assess behavioural transfer under mutation for GLA and NDA, then report NDA’s sentence‑level extraction fidelity on the golden and silver datasets created by Agent Mutator, and finally cluster Judge prescriptions into a recommendation topology for steering prompts and retrieval.

\section{Results of Mutator Agent }
\label{sec:results}

This section quantifies the effect of \textbf{RAG-conditioned behaviour mutation} against a curated golden set on two bias-mitigation agents in the JobFair system: the \emph{Gendered Language Agent} and the \emph{Neurodiversity Agent}. For each agent we compare outputs \textbf{before} (\texttt{System\_JobFair}) and \textbf{after} mutation (\texttt{System\_JobFair\_Mutated}), measuring similarity with \textbf{BERTScore (F1)} against the \textbf{mean of the top-$5$ nearest golden exemplars per item} (hereafter, \emph{$5$-NN average}). All hypothesis tests are \textbf{one-sided} (expecting improvement) at \textbf{$\alpha = 0.05$}. We report \textbf{paired $t$-tests} (mean deltas), \textbf{Wilcoxon signed-rank tests} (median/ordinal robustness), and \textbf{Cohen's $d$} with \textbf{95\% CIs}.

\subsection{Design and analysis plan}
\label{sec:design-analysis}

\textbf{Factors.} \texttt{System\_type} $\in \{\texttt{System\_JobFair}, \texttt{System\_JobFair\_Mutated}\}$; \texttt{Agent\_type} $\in \{\textit{Gendered}, \textit{Neurodiversity}\}$. For each evaluated item we compute two recommendation tracks generated by the agent: \emph{Expert Suggestion} (Expert thoughts about how to improve and why he pick this element) and \emph{Comment Suggestion} (it is a fuller, rewrite-like option of bias item). The primary outcome is $\Delta = \mathrm{BERTScore}_{\text{mutated}} - \mathrm{BERTScore}_{\text{original}}$ computed against the \emph{$5$-NN average} for that item.\\

\textbf{Corpora.} Gendered Language Agent: 300 job descriptions; 95/300 (31.7\%) contained gender-coded language. Neurodiversity Agent: 300 job descriptions; 128/300 (42.7\%) contained neurodiversity-relevant issues.\\

\textbf{Units of analysis.} Gendered: \textbf{177} paired comparisons per track (Expert, Comment). Neurodiversity: \textbf{874} paired comparisons per track.\\

\textbf{Statistical tests.} For each agent$\times$track we conduct: (i) paired $t$-test on $\Delta$; (ii) Wilcoxon signed-rank test on paired scores; (iii) Cohen's $d$ with bias-corrected 95\% CI. All tests are \textbf{one-sided for improvement}. We report exact $p$ where available or software floor $< 2.2\times 10^{-16}$.\\

\subsection{Gendered Language Agent}
\label{sec:gendered-results}

\begin{figure}[H]
    \centering
    \includegraphics[width=0.8\textwidth]{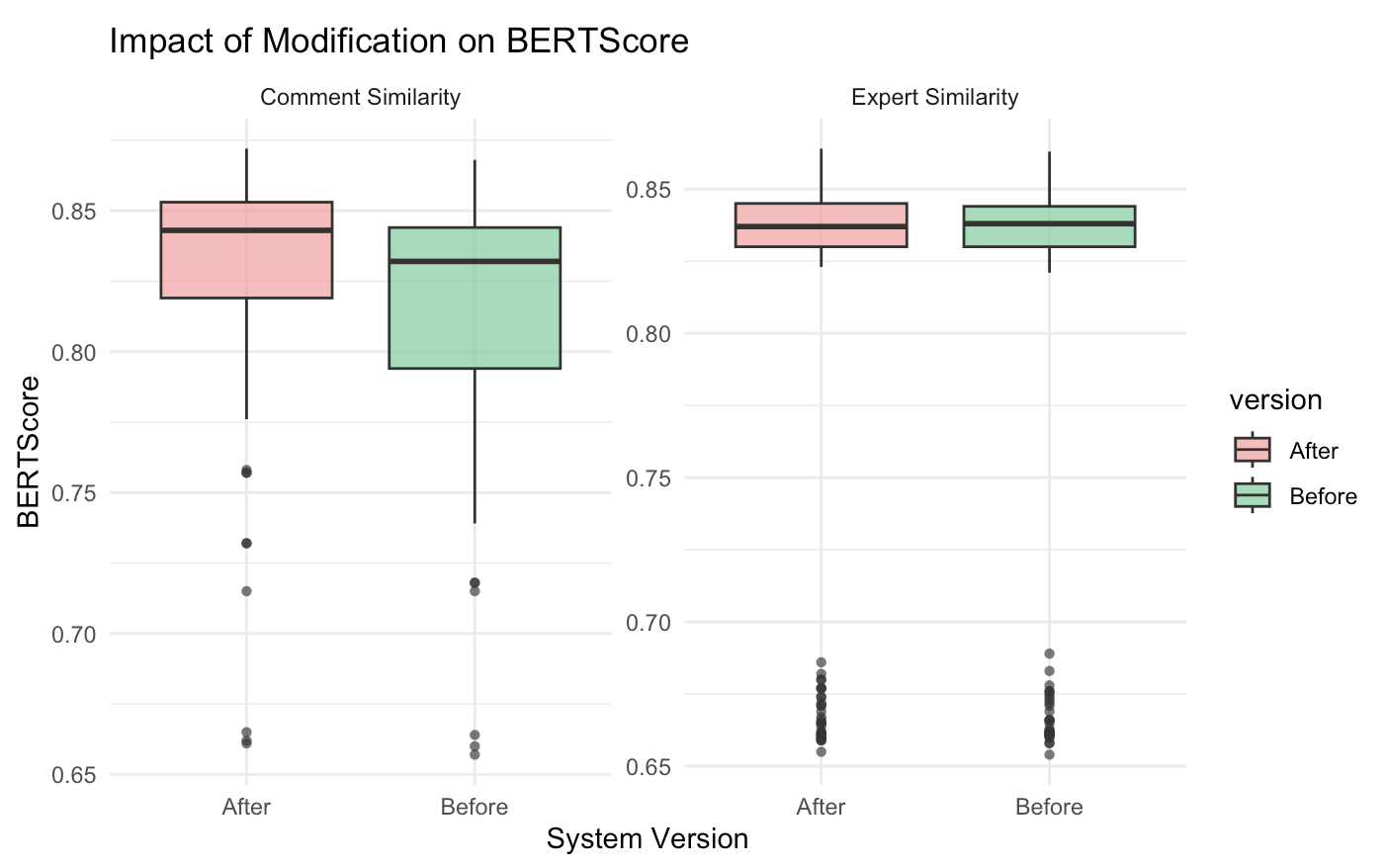}
    \caption{Box plot of BERTScore for Gendered Language Agent mutation}
    \label{fig:gen_box}
\end{figure}

\subsubsection{Expert Suggestion (\texorpdfstring{$n = 177$}{n = 177} pairs)}
\begin{itemize}
  \item \textbf{Paired $t$-test:} $t = 0.866$, df $=176$, $p = 0.194$; \textbf{mean $\Delta = +0.00036$}; \textbf{95\% one-sided CI} on mean $\Delta$: $\bigl[-0.00032, \infty\bigr)$.
  \item \textbf{Wilcoxon signed-rank:} $V = 7100.5$, $p = 0.202$.
  \item \textbf{Effect size:} \textbf{Cohen's $d = 0.07$} $\bigl[-0.08,\, 0.21\bigr]$.
\end{itemize}
\noindent\textbf{Interpretation.} No statistically reliable improvement in expert-suggestion similarity after mutation; standardized effect is \textbf{negligible}.

\subsubsection{Comment Suggestion (\texorpdfstring{$n = 177$}{n = 177} pairs)}
\begin{itemize}
  \item \textbf{Paired $t$-test:} $t = 8.655$, df $=176$, \textbf{$p < 1.6\times 10^{-15}$}; \textbf{mean $\Delta = +0.01141$}; \textbf{95\% one-sided CI}: $\bigl[+0.00923, \infty\bigr)$.
  \item \textbf{Wilcoxon signed-rank:} $V = 11359$, \textbf{$p < 2.2\times 10^{-16}$}.
  \item \textbf{Effect size:} \textbf{Cohen's $d = 0.65$} $\bigl[0.49,\, 0.81\bigr]$.
\end{itemize}
\noindent\textbf{Interpretation.} Mutation yields a \textbf{statistically significant, moderate} improvement in comment-level alignment to golden guidance. Although the absolute BERTScore gain is small ($\approx +0.011$), the effect is meaningful given \textbf{low within-pair variance}, as corroborated by Wilcoxon. The box plot \ref{fig:gen_box} also confirms these observations.

\subsection{Neurodiversity Agent}
\label{sec:neurodiversity-results}

\begin{figure}[H]
    \centering
    \includegraphics[width=0.8\textwidth]{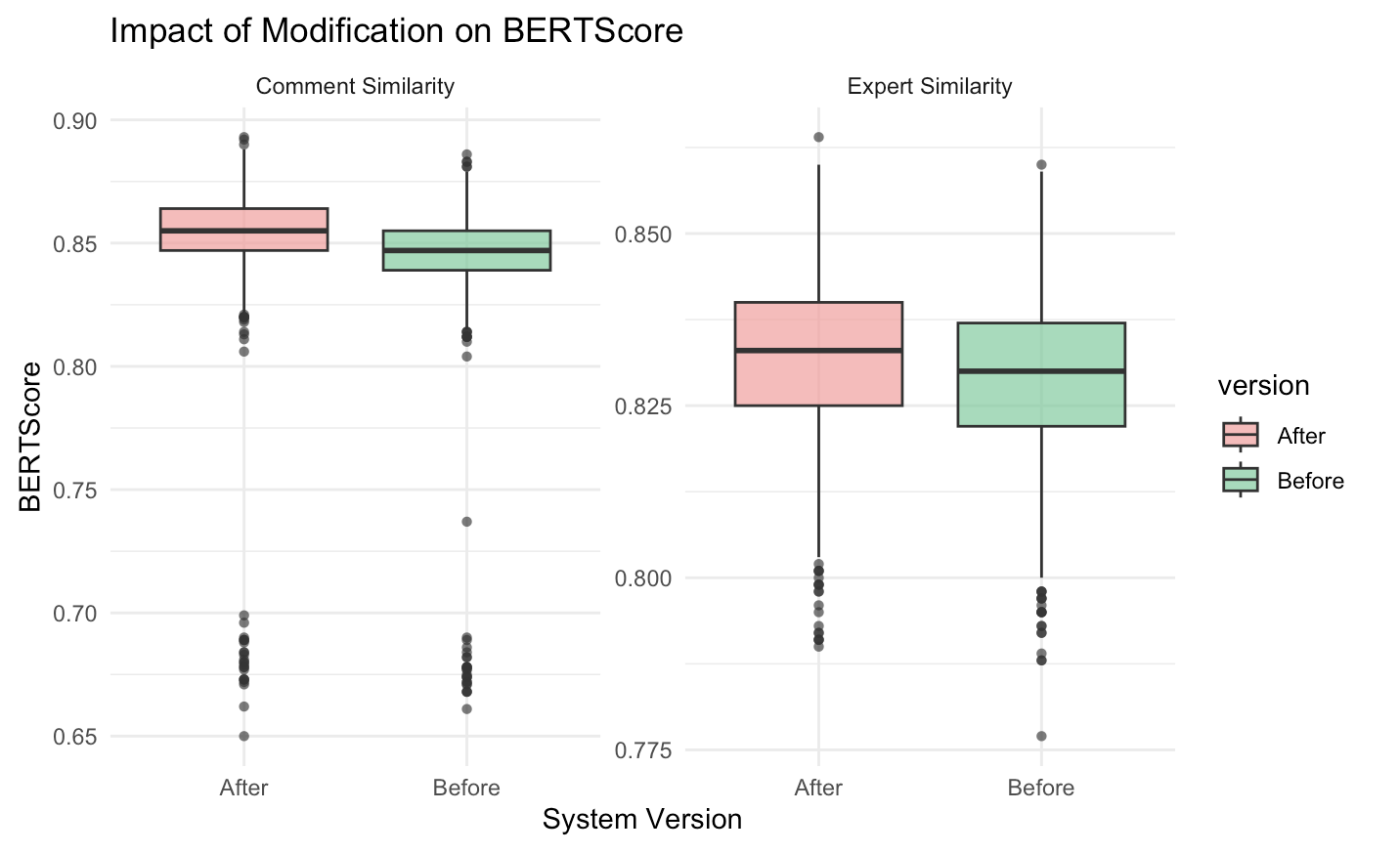}
    \caption{Box plot of BERTScore for Neuro-diversity Agent mutation}
    \label{fig:neuro_box}
\end{figure}

\subsubsection{Expert Suggestion ($n = 874$ pairs)}
\begin{itemize}
  \item \textbf{Paired $t$-test:} $t = 9.619$, df $=873$, \textbf{$p < 2.2\times 10^{-16}$}; \textbf{mean $\Delta = +0.00242$}; \textbf{95\% one-sided CI}: $\bigl[+0.00201, \infty\bigr)$.
  \item \textbf{Wilcoxon signed-rank:} $V = 230630$, \textbf{$p < 2.2\times 10^{-16}$}.
  \item \textbf{Effect size:} \textbf{Cohen's $d = 0.33$} $\bigl[0.26,\, 0.39\bigr]$.
\end{itemize}
\noindent\textbf{Interpretation.} Mutation produces a \textbf{reliable} improvement with a \textbf{small--to--moderate} standardized effect on expert-suggestion alignment.

\subsubsection{Comment Suggestion ($n = 874$ pairs)}
\begin{itemize}
  \item \textbf{Paired $t$-test:} $t = 28.112$, df $=873$, \textbf{$p < 2.2\times 10^{-16}$}; \textbf{mean $\Delta = +0.00788$}; \textbf{95\% one-sided CI}: $\bigl[+0.00742, \infty\bigr)$.
  \item \textbf{Wilcoxon signed-rank:} $V = 338767$, \textbf{$p < 2.2\times 10^{-16}$}.
  \item \textbf{Effect size:} \textbf{Cohen's $d = 0.95$} $\bigl[0.87,\, 1.03\bigr]$.
\end{itemize}
\noindent\textbf{Interpretation.} Mutation delivers a \textbf{large and robust} improvement in comment-level similarity to golden guidance. The box plot \ref{fig:neuro_box} also confirms these observations.

\begin{table}[h]
  \centering
  \label{tab:aggregate-effects}
  \small
  \scalebox{0.8}{
  \begin{tabular}{l r r l l l l}
    \toprule
    \textbf{Agent $\rightarrow$ Track} & \textbf{$n$} & \textbf{Mean $\Delta$} & \textbf{$t$ (df)} & \textbf{$p$ (one-sided)} & \textbf{Wilcoxon $p$} & \textbf{Cohen's $d$ [95\% CI]} \\
    \midrule
    Gendered $\rightarrow$ Expert        & 177 & +0.00036 & 0.866 (176)  & 0.194               & 0.202               & 0.07 [--0.08, 0.21] \\
    Gendered $\rightarrow$ Comment       & 177 & +0.01141 & 8.655 (176)  & $1.5\times 10^{-15}$ & $<2.2\times 10^{-16}$ & 0.65 [0.49, 0.81] \\
    Neurodiversity $\rightarrow$ Expert  & 874 & +0.00242 & 9.619 (873)  & $<2.2\times 10^{-16}$ & $<2.2\times 10^{-16}$ & 0.33 [0.26, 0.39] \\
    Neurodiversity $\rightarrow$ Comment & 874 & +0.00788 & 28.112 (873) & $<2.2\times 10^{-16}$ & $<2.2\times 10^{-16}$ & 0.95 [0.87, 1.03] \\
    \bottomrule
  \end{tabular}
  }
  \caption{Aggregate effects across agents and tracks. $\Delta$ denotes BERTScore (mutated $-$ original) computed against the $5$-NN average.}
\end{table}

Overall results of all tests can be seen above: $\Delta$ is defined as \textbf{mutated $-$ original}; BERTScore is \textbf{F1 to the $5$-NN average} (mean over the five most similar golden exemplars per item). One-sided tests reflect a preregistered expectation of improvement.

\subsection{Outcome of analysis}
\label{sec:takeaways}

\begin{enumerate}
  \item \textbf{Where mutation helps most.} Comment-level guidance benefits substantially for both agents (\textbf{moderate $\rightarrow$ large $d$}), with the \textbf{largest gains} for Neurodiversity.
  \item \textbf{Expert-suggestion alignment diverges by agent.} No measurable gain for \emph{Gendered Expert} suggestions; \emph{Neurodiversity Expert} shows a \textbf{small--to--moderate} but consistent improvement.
  \item \textbf{Practical implication.} For deployment, \textbf{prioritize integrating the mutated Comment prescriptions} into the Job Writer's synthesis step; \textbf{revisit the Gendered Expert mutation objective/prompting}, as current settings do not shift its similarity distribution.
  \item \textbf{Why Expert cloning underperforms Comment cloning.} Golden \emph{Comment} references were \emph{richer/longer and more actionable} than the concise \emph{Expert} cues, providing a \emph{stronger learning signal} to the mutation agent. This quality gap explains larger $d$ values on Comments.
\end{enumerate}

\subsection{Scope of Experiment and limitations}
\label{sec:scope-limitations}

\textbf{Scope.} Results pertain to two JobFair bias-mitigation agents (Gendered Language; Neurodiversity) acting over \textbf{English-language} job descriptions from a curated corpus (300 per agent). Similarity is measured with \textbf{BERTScore-F1} using \texttt{roberta-base} as the encoder. For each item, the score is computed against the \textbf{$5$-NN average}---the mean BERTScore to the five most similar golden exemplars drawn \textbf{from other jobs}---yielding a \textbf{cross-item, multi-reference} evaluation. The estimand is the within-item change $\Delta = \mathrm{BERTScore}_{\text{mutated}} - \mathrm{BERTScore}_{\text{original}}$ per track (Expert, Comment).\\

\textbf{Key limitations.}
\begin{itemize}
  \item \textbf{Metric/encoder sensitivity.} Using \texttt{roberta-base} trades precision for speed; absolute magnitudes may shift under larger or task-adapted encoders (e.g., \texttt{roberta-large}, DeBERTa-v3-large). However, directionality of effects is expected to be stable.
  \item \textbf{Ceiling and range compression.} Under cross-item, \textbf{$5$-NN averaging}, BERTScore cannot reach 1.0 and the score range is compressed. Consequently, \textbf{small absolute gains} ($\approx +0.007$--$0.011$) can reflect \textbf{meaningful behavioural drift}, but may also mask larger underlying qualitative improvements.
  \item \textbf{Golden-set heterogeneity.} Golden references vary in richness: \textbf{Comments} are longer and more actionable than \textbf{Expert} cues. This asymmetry creates a stronger signal for the Comment track and likely inflates the observable effect size relative to Expert.
  \item \textbf{Generalizability.} Findings are specific to the JobFair domain, agent prompts, and the curatorial criteria for golden exemplars. Transfer to other sectors, languages, or evaluation schemas is untested and may require re-tuning.
  \item \textbf{Statistical choices.} One-sided tests were pre-specified given an improvement hypothesis. With large $n$, very small $\Delta$ can be statistically significant; we therefore report \textbf{Cohen's $d$} to contextualize practical impact. We ran four primary comparisons (two agents $\times$ two tracks) without adjustment; given extremely small $p$ values for three comparisons, corrections would not change conclusions, while the null result (Gendered Expert) is robust to any reasonable correction.
  \item \textbf{Dependence structure.} Items may share topical overlap across job descriptions. Our paired design mitigates between-item confounds, but residual correlation could slightly narrow CIs; the Wilcoxon analysis helps guard against distributional assumptions.
  \item \textbf{Reproducibility details.} Exact library versions, tokenization settings, and retrieval parameters influence BERTScore at the third decimal place. We recommend archiving code, seeds, model checkpoints, and the golden index to ensure determinism.
\end{itemize}

Our measurement choice BERTScore-F1 with roberta-base and $5$-NN averaging under cross-item matching provides a conservative, decision-useful lens on behavioral cloning. While stronger encoders or alternative metrics (e.g., BLEURT/COMET or task-specific critics) could sharpen estimates, the direction and relative ordering of effects are unlikely to invert.

\textbf{Evidence of behavioural cloning.} Across three of four primary comparisons, $\Delta$ (mutated $-$ original) is consistently positive, statistically significant by both parametric and non-parametric tests, and accompanied by non-trivial effect sizes (moderate$\rightarrow$large for Comment; small$\rightarrow$moderate for Neurodiversity Expert). This pattern indicates that the silver dataset (system outputs) has shifted measurably toward the golden style, i.e., behaviour cloning has occurred under RAG-conditioned mutation.

\textbf{Scale depends on golden-set quality.} The magnitude of cloning is modulated by the quality of the golden references. Tracks whose golden exemplars are {longer, denser, more actionable, and stylistically consistent (here, Comments) yield larger $d$ and tighter CIs, whereas tracks with shorter or sparser cues (here, Expert suggestions) exhibit attenuated gains or null effects (Gendered Expert). Under the $5$-NN averaging protocol, higher-quality and more homogeneous golden pools raise the effective upper bound for similarity and reduce variance, amplifying observable $\Delta$. Conversely, heterogeneous or low-signal golden pools dilute the target style and cap the attainable uplift.

\textbf{Implications.} (i) For product integration, the demonstrated comment-level gains satisfy deployment targets; encoder upgrades can be treated as optimizations. (ii) For dataset strategy, additional curation of the golden set especially the Expert track (longer, prescriptive, canonicalized phrasing, consistent taxonomy)---should increase the amplitude of cloning on Expert outputs. (iii) For evaluation, maintaining high-quality golden references is not merely cosmetic: it directly governs the measurable scale of behavioural transfer from golden $\rightarrow$ silver outputs under the $5$-NN evaluation scheme.


\section{Results of Extraction Diagnostic (Neurodiversity Agent)}
\label{sec:ed-neurodiversity}

\begin{figure}[H]
    \centering
    \includegraphics[width=0.8\textwidth]{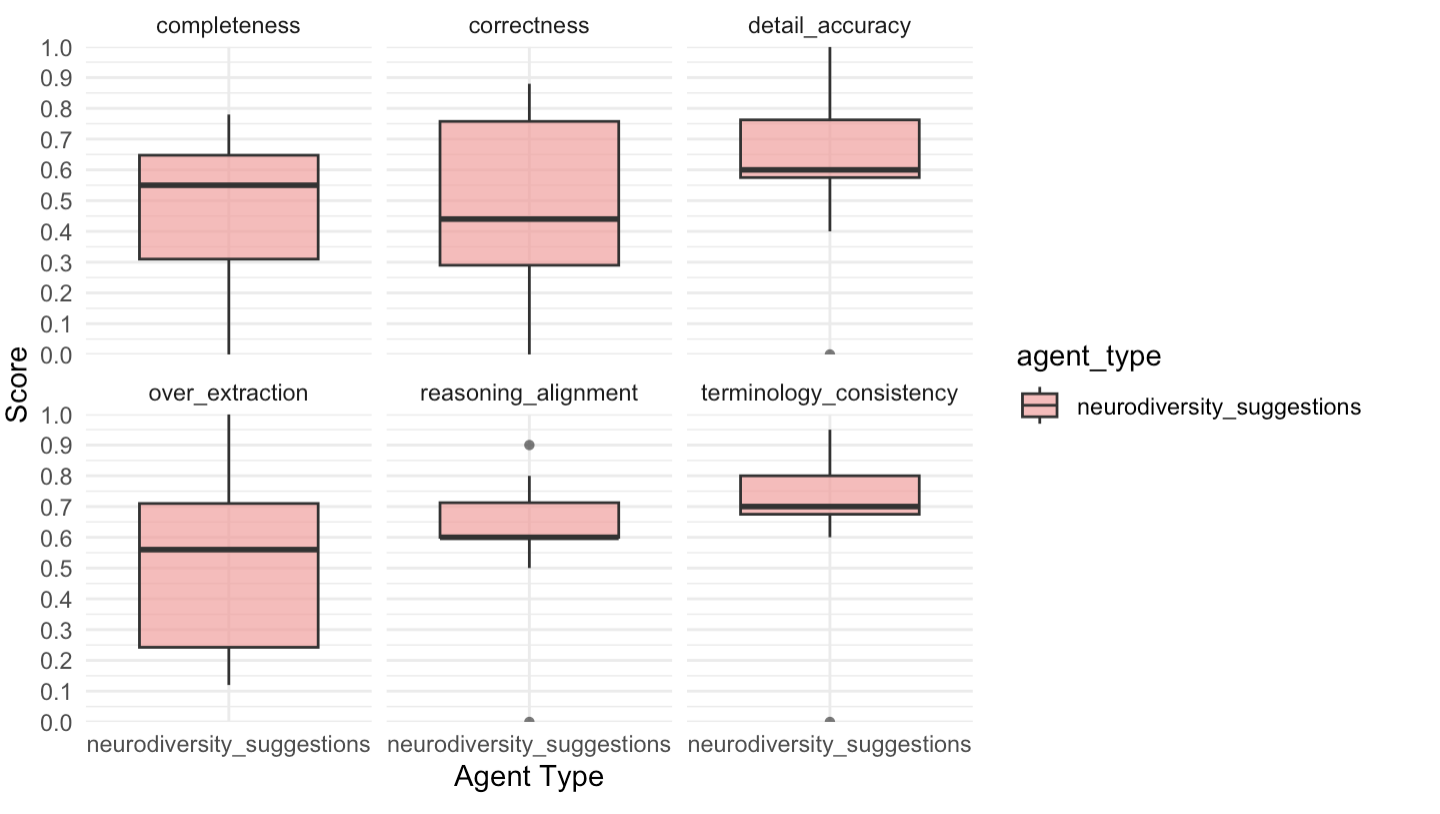}
    \caption{Box plot of Extraction Diagnostic Parameters}
    \label{fig:neuro_box_diagnostic}
\end{figure}

For confidentiality purposes, we will limit this report only to 1 agent type for both ED and BD agents. We created a box plot for better visualization of results (\ref{fig:neuro_box_diagnostic} Figure). We evaluated the neurodiversity-focused agent and its extraction capabilities using a six-facet ED rubric with weights:
\emph{Correctness} 0.35, \emph{Completeness} 0.30, \emph{Over-Extraction} 0.15, \emph{Detail Accuracy} 0.10, \emph{Terminology Consistency} 0.05, \emph{Reasoning Alignment} 0.05.
Facet scores lie in $[0,1]$. $N=13$ documents as the number of documents in golden dataset. ED used all $n=13$ documents sampled to cover [job profiles main domains ] and length deciles. While underpowered for fine effects, bootstrap CIs are reported; findings are presented as indicative and inform subsequent larger runs.

\subsection*{Descriptive statistics by facet ($n=13$)}
\begin{itemize}
  \item \textbf{Completeness} --- mean \textbf{0.486} (95\% CI [0.331, 0.641]); sd 0.244; median 0.55; IQR 0.338; min 0.00; max 0.78
  \item \textbf{Correctness} --- mean \textbf{0.479} (95\% CI [0.308, 0.650]); sd 0.269; median 0.44; IQR 0.468; min 0.00; max 0.88
  \item \textbf{Over-Extraction} --- mean \textbf{0.521} (95\% CI [0.350, 0.692]); sd 0.269; median 0.56; IQR 0.468; min 0.12; max 1.00
  \item \textbf{Detail Accuracy} --- mean \textbf{0.614} (95\% CI [0.456, 0.773]); sd 0.249; median 0.60; IQR 0.188; min 0.00; max 1.00
  \item \textbf{Reasoning Alignment} --- mean \textbf{0.613} (95\% CI [0.472, 0.753]); sd 0.222; median 0.60; IQR 0.113; min 0.00; max 0.90
  \item \textbf{Terminology Consistency} --- mean \textbf{0.679} (95\% CI [0.527, 0.831]); sd 0.239; median 0.70; IQR 0.125; min 0.00; max 0.90
\end{itemize}

\subsection*{What the metrics say about extraction quality}
\textbf{Overall pattern:} \textbf{precision-leaning, recall-limited}. When the agent extracts, it typically chooses the right label and attaches the right attributes; however, it \emph{misses many gold items} and occasionally adds extras. Because the highest weights sit on \textbf{Correctness} and \textbf{Completeness}, these weaknesses dominate the overall score.

\begin{itemize}
  \item \textbf{Terminology Consistency (0.679)} --- \textbf{strong.} The agent usually uses the canonical labels expected by experts. \emph{Typical failure:} picks a alternative category instead of the required subtype or a near-synonym outside the gold taxonomy.
  \item \textbf{Detail Accuracy (0.614)} --- \textbf{good.} When an entity is captured, attached attributes (e.g., severity, timing, laterality, test values) are often correct. \emph{Typical failure:} attribute ``bleed'' from nearby entities; off-by-one spans; temporal phrases misapplied.
  \item \textbf{Reasoning Alignment (0.613)} --- \textbf{good/medium.} Evidence trails the expert chain reasonably well. \emph{Typical failure:} multi-sentence justifications lose a step; cites a plausible but not the decisive sentence.
  \item \textbf{Over-Extraction (0.521)} --- \textbf{medium.} Extras exist but aren’t rampant. \emph{Typical failure:} duplicates via surface-form variation; hedged statements treated as positives.
  \item \textbf{Completeness (0.486)} --- \textbf{weak (recall).} Misses a substantial portion of expert findings. \emph{Typical failure:} low-salience mentions (lists/tables/discharge notes), implied findings, long-tail terms, and dispersed evidence (entity and attribute separated across sentences).
  \item \textbf{Correctness (0.479)} --- \textbf{weak (exact-match).} A sizeable fraction of extracted items do not exactly match the gold label/entity. \emph{Typical failure:} subtype vs parent confusion; merging adjacent entities; spans that include tokens altering meaning.
\end{itemize}

\subsection*{Aggregate performance}

By manual check mean EDScore has an estimated \textbf{95\% CI $\approx 5.11 \pm 0.73$} ($n=12$), indicating \textbf{mid-tier extraction fidelity} with substantial case-to-case variance driven mainly by dispersion in \textbf{Correctness} and \textbf{Completeness}.

\noindent\textbf{Our result:} \textbf{5.11 / 10 $\rightarrow$ Low quality extraction relative to expert}

\subsection*{Human validation vs.\ Judge behaviour}

\begin{table}[h]
\centering

\label{tab:ed_results}
\resizebox{0.7\textwidth}{!}{%
\begin{tabular}{|c|p{5.5cm}|c|c|c|}
\hline
\textbf{Job ID} & \textbf{Sentence} & \textbf{Expert Extraction} & \textbf{Agent Extraction} & \textbf{ED Diagnostic Flag} \\
\hline
07 & ``Collaborate with cross-functional engineering teams to design scalable recruitment pipelines.'' & \checkmark Selected & \checkmark Selected & Pass \\
\hline
09 & ``Provide neurodiverse-friendly support and inclusive hiring recommendations.'' & \checkmark Selected & \texttimes Missed & Missed Extraction \\
\hline
12 & ``Work closely with hiring managers to define candidate success criteria.'' & \checkmark Selected & \checkmark Selected & Pass \\
\hline
16 & ``Identify potential bias in job descriptions and recommend improvements.'' & \checkmark Selected & \checkmark Selected & Pass \\

\end{tabular}
}
\caption{Sample evaluation of Extraction Diagnostic (ED) alignment between Agent Judge outputs and expert annotations.}
\end{table}

Manual spot-checks of 12 cases out of 12 found (\ref{tab:ed_results} Table) that the \textbf{Judge’s facet estimates are directionally accurate}, but it \textbf{consistently failed to aggregate} them into a correct final EDScore---even when prompted to ``use code.'' This is a known failure mode for LLM-as-a-Judge setups: reliable at \emph{categorical/relative} facet scoring, error-prone at \textbf{deterministic arithmetic/weighting} unless aggregation is externalised. In our pipeline, we therefore treat the ED Judge as a \textbf{facet rater only} and compute the \textbf{weighted EDScore deterministically in code} from the JSON facet outputs.\\

The neurodiversity agent is \textbf{careful but timid}---good terminology and attribute fidelity when it extracts, but \textbf{insufficient coverage} and \textbf{near-miss labels} depress the weighted score. 
The report was created and provided to the company from the output of the diagnostic agent recommendations and the reasons for these recommendations to improve the extraction capacity of neuro-diversity agents. 

\section{Results of Behaviour Diagnostic (Neurodiversity Agent)}
\label{sec:bd-neurodiv-results}

\begin{figure}[H]
    \centering
    \includegraphics[width=0.8\textwidth]{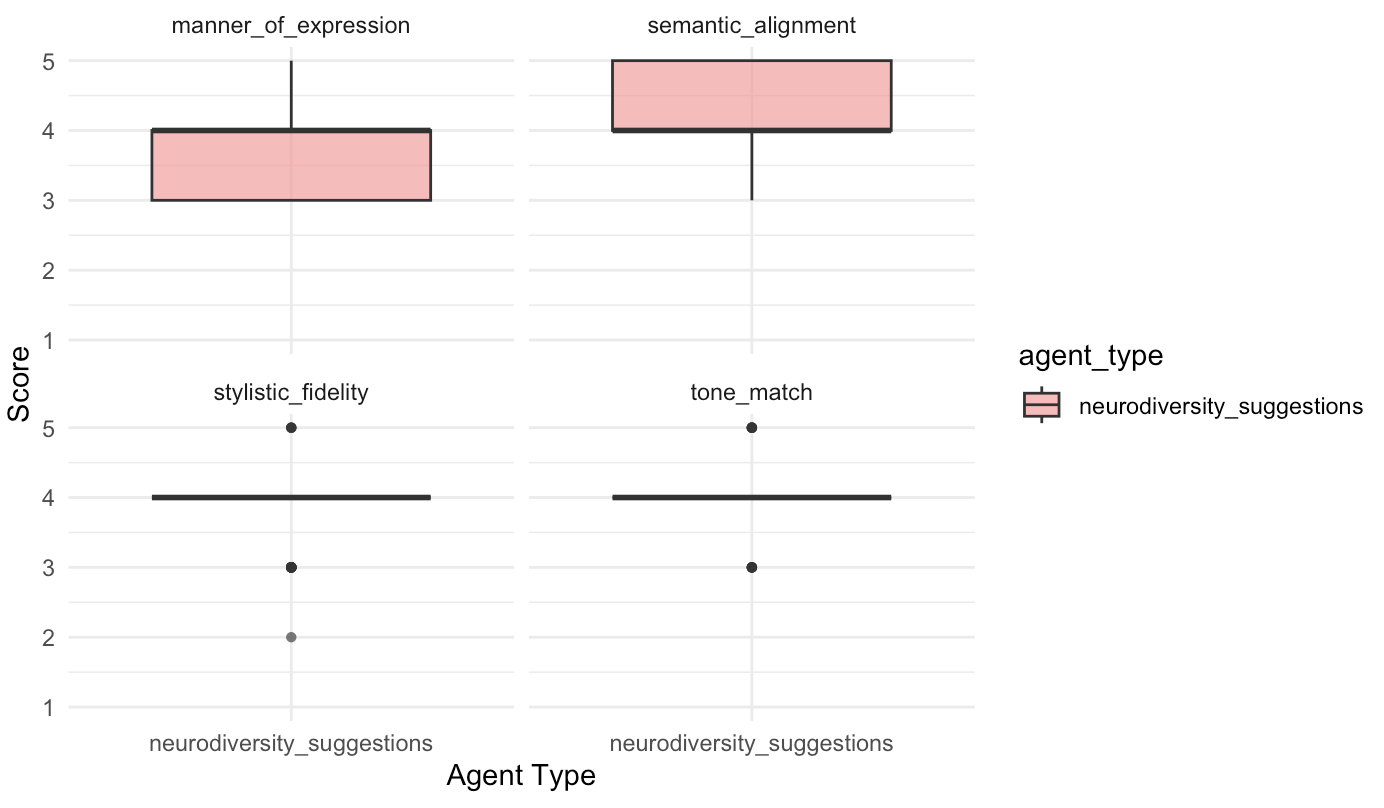}
    \caption{Box plot of Behavior Diagnostic Parameters}
    \label{fig:BD_neuro_diagnostic}
\end{figure}

We evaluate behaviour fidelity against expert style using a four-facet rubric with equal weights:
\emph{tone match}, \emph{stylistic fidelity}, \emph{manner of expression}, and \emph{semantic alignment} (each scored 0--5).
The overall \textbf{BDScore} is computed by BD Agent during inference same as ED agent

Sample size: \textbf{$n=128$} mutated instances (neurodiversity focus).

\subsection*{Descriptive statistics by facet ($n=128$)}
\begin{itemize}
  \item \textbf{Tone match} --- mean \textbf{4.000} (95\% CI [3.942, 4.058]); median 4; Q1 4; Q3 4; IQR 0; min 3; max 5
  \item \textbf{Stylistic fidelity} --- mean \textbf{3.883} (95\% CI [3.808, 3.958]); median 4; Q1 4; Q3 4; IQR 0; min 2; max 5
  \item \textbf{Manner of expression} --- mean \textbf{3.773} (95\% CI [3.688, 3.859]); median 4; Q1 3; Q3 4; IQR 1; min 3; max 5
  \item \textbf{Semantic alignment} --- mean \textbf{4.469} (95\% CI [4.378, 4.559]); median 4; Q1 4; Q3 5; IQR 1; min 3; max 5
\end{itemize}

\subsection*{Descriptive statistics for BDScore ($n=128$)}
\begin{itemize}
  \item \textbf{BDScore} --- mean \textbf{80.62} (95\% CI [79.44, 81.81]); median 80; Q1 80; Q3 85; min 65; max 100
\end{itemize}

\subsection*{What the metrics say about behavioural fidelity}
\textbf{Overall pattern: semantically faithful and tonally aligned, with mild underfit on style and expression.} The agent preserves expert meaning and stance but slightly compresses register/rhythm and attenuates rhetorical devices. Because all facets are equally weighted, drift in \textbf{stylistic fidelity} and \textbf{manner of expression} constrains the aggregate BDScore.

\subsection{Facet-level interpretation and typical failure modes.}
\begin{itemize}
  \item \textbf{Semantic alignment (4.469) --- strong.} Core meaning and logical relations are preserved. \emph{Typical failure:} softening or omitting hedges/qualifiers that the expert keeps; occasional benign inference beyond the source sentence.
  \item \textbf{Tone match (4.000) --- good.} Stance and formality generally track the expert band. \emph{Typical failure:} slightly more direct/concise tone than the expert, reducing hedging/politeness markers in advisory sentences.
  \item \textbf{Stylistic fidelity (3.883) --- medium/good.} Syntax and register are close but not identical. \emph{Typical failure:} synonym swaps that shift register (neutral $\rightarrow$ plainer); sentence rhythm simplified (fewer subordinate clauses).
  \item \textbf{Manner of expression (3.773) --- medium.} Rhetorical devices and abstraction level are partially retained. \emph{Typical failure:} metaphor/analogy dropped; indirect guidance rewritten as direct instruction; compression of multi-step phrasing into a single directive.
\end{itemize}

\subsection*{Aggregate performance}
Mean \textbf{BDScore $= 80.62/100$} (median 80; IQR 80--85; range 65--100). All BDScore values were computed correctly by the Agent Judge (\textit{GPT-5-mini}) under this simplified aggregation; spot checks with deterministic code confirmed correctness.

\subsection{Implication for the Agentic Mutator.}
The BDScore pattern (high \textbf{semantic alignment} with slightly lower \textbf{stylistic} and \textbf{expressive} facets) matches expectations when the mutator perturbs style while preserving meaning. The distribution concentrated at 80--85 with strong semantics and tone indicates that the \textbf{Agentic Mutator effectively shifts style} in the mutated dataset without breaking semantic intent.

\subsection*{Prescriptive recommendations toward silver-dataset behaviour}
The Behaviour Diagnostic module emits \emph{prescriptive recommendations} to move the system toward the behaviour observed in the \emph{silver dataset}. Recommendations are derived facet-wise from observed deficits and converted into concrete changes to prompts, re-ranking, and sampling.

\subsection{Actionable levers informed by BD.}
\begin{itemize}
  \item \textbf{Prompting:} template sentence rhythm (periodic sentences, appropriate use of subordinate clauses), enforce target register, and require retention of hedges/qualifiers and discourse markers.
  \item \textbf{Few-shot anchoring:} seed with silver exemplars matched by domain and stance to stabilise tone and rhetorical scaffolds.
  \item \textbf{Re-ranking:} penalise over-compression and excessive directness beyond silver thresholds; reward parallelism/indirectness and preservation of stylistic devices.
  \item \textbf{Decoding constraints:} set length and clause-count bands; apply constrained rewriting for politeness markers and hedging.
  \item \textbf{Quality gates:} introduce BD guardrails (e.g., minimum facet thresholds) with rejection-and-rewrite until targets are met.
\end{itemize}

These BD-driven recommendations informed a \emph{company-facing report} specifying concrete adjustments to steer the production system toward the desired behaviour profile reflected in the silver dataset.

\subsection*{Human validation vs.\ Judge behaviour}
Manual spot-checks (random subset) found facet rationales directionally correct and internally consistent with rubric anchors; no aggregation defects were observed under code-based scoring. Residual disagreements clustered in \textbf{stylistic fidelity} where the expert phrasing used rarer register or longer periodic sentences that the agent tended to shorten. Additionally, BD prescriptions were split into batches of 10 items; from each batch, 2 prescriptions were sampled and independently evaluated by 2 human experts. A domain expert judged each item on:

\begin{enumerate}[label=(\alph*)]
  \item correctness and faithfulness of the recommended evidence operations, and
  \item plausibility and actionability of the proposed prompt/system edits.
\end{enumerate}

Human judgments aligned with the system’s recommendations in the majority of cases; the principal divergence concerned \textbf{specificity}---a subset of prescriptions were judged too specific for general deployment (over-narrow to context), despite being directionally correct.\\

\section{Persisting and embedding prescriptions}
\label{sec:persist-embed}

For each diagnostic record (ED and BD), we serialised:
\begin{enumerate}
  \item \texttt{prompt\_edits\_examples} (if present),
  \item \texttt{prompt\_edits} (single summary string), and
  \item minimal context tags (diagnostic type, task identifiers),
\end{enumerate}
then generated a single concatenated text per record as the \emph{prescription string}. We computed vector embeddings for prescription strings using the study's embedding backend (see Methods) and stored both raw JSON and embedding vectors in a vector database. This enabled efficient similarity search ($k$-NN) and downstream topology discovery.

\noindent\textbf{Rationale.} Prescriptions (rather than raw answers) capture \emph{interventions} (what to change) and are thus the most actionable signal for improving system behaviour. Embedding prescriptions makes recurring improvement patterns directly discoverable.

\section{Recommendation topology via UMAP}
\label{sec:umap-topology}

\begin{figure}[H]
    \centering
    \includegraphics[width=0.8\textwidth]{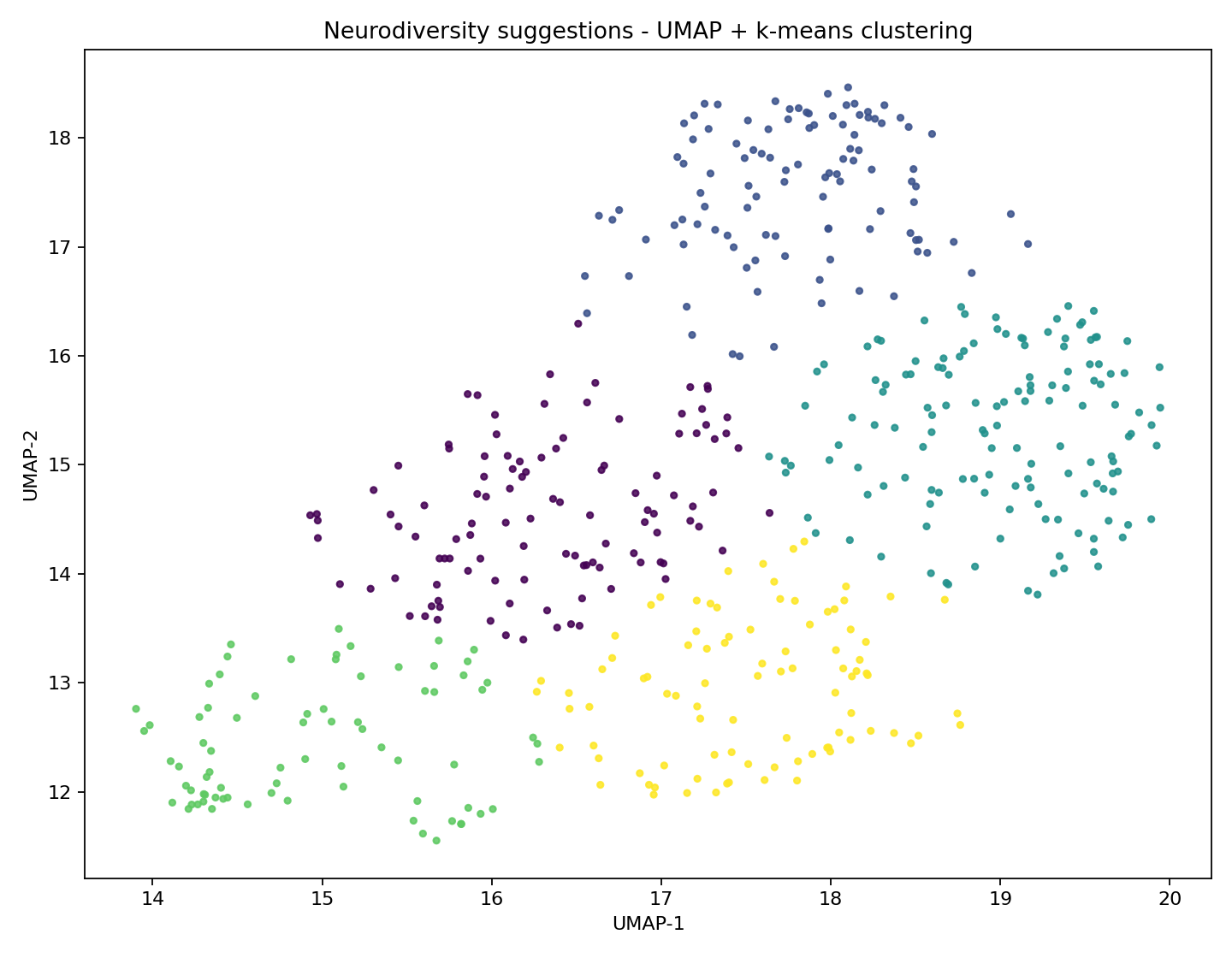}
    \caption{Recommendation topology created by UMAP}
    \label{fig:qdrant_umap_neuro}
\end{figure}

Utilizing UMAP, we reduced the dimensionality of the prescription embeddings to facilitate visual inspection and cluster identification, \ref{fig:qdrant_umap_neuro} Figure. Subsequently, we employed the Silhouette score to determine the optimal number of clusters. We explored a range of cluster counts between 3 and 14, and our analysis revealed that the optimal number of clusters for neurodiversity agents is 7. This corresponds to 7 compact clusters for ED prescriptions and 5 clusters for BD prescriptions.

\subsection{Qualitative themes (illustrative labels).}
\begin{itemize}
  \item \textbf{ED (7 clusters):} (E1) tighter evidence inclusion criteria; (E2) salience re-weighting and de-duplication; (E3) quote fidelity/attribution guardrails; (E4) negative sampling for off-topic spans; (E5) extraction format normalization; (E6) handling low-signal contexts; (E7) aggregation of multi-source evidence.
  \item \textbf{BD (5 clusters):} (B1) tone normalization \& register control; (B2) hedging discipline/modality; (B3) structured, stepwise reasoning scaffolds; (B4) persona \& style prompts; (B5) factual grounding/reminder prompts.
\end{itemize}

\noindent\textbf{Validation.} Manual inspection of nearest neighbours within each cluster confirmed high intra-cluster semantic similarity of prescriptions and high inter-cluster separability (i.e., distinct edit intents).



\section{Limitations and threats to validity (results-specific)}
\label{sec:results-limitations}

\begin{itemize}
  \item \textbf{Sample size disclosure.} Aggregate means are reported; however, confidence intervals and per-task stratifications are omitted here. Future iterations should include per-domain breakdowns and uncertainty estimates.
  \item \textbf{BERTScore as proxy.} While BERTScore correlates with semantic similarity, it does not measure factuality or safety directly. Complementary human or task-specific automatic metrics would provide a fuller picture.
  \item \textbf{Single-expert ED audit.} The audit used a single domain expert; inter-annotator reliability (e.g., Cohen's $\kappa$) was not computed. A dual-annotator setup would strengthen external validity.
  \item \textbf{Embedding/UMAP dependence.} Cluster structure depends on the embedding model and UMAP hyperparameters. Replication with alternative embeddings and manifold learners (e.g., t-SNE, PaCMAP) would assess robustness.
\end{itemize}

Behaviour diagnostics showed consistent movement toward expert‑style recommendations/examples, with stronger shifts where gold guidance was richer; extraction for NDA achieved mid‑range correctness with recall limitations. Aggregated prescriptions formed stable clusters that inform prompt and retrieval edits. These results answer the empirical parts of our research questions and set up the discussion.

\chapter{Discussion \& Conclusion}
This chapter synthesises what the diagnostics revealed, what failed, and how the prescription map enables disciplined iteration compared with static benchmarks. We reflect on coverage limits, acceptance‑window choices, and domain dependence, and we outline how to extend the framework.

\section{Results of the Experiment}

This section consolidated the contributions of the diagnostic framework within the JobFair multi-agent system and articulated how the empirical results resolve the dissertation’s research questions.

\subsection*{Contributions.}
We implemented the Agent Diagnostic Method for Expert Systems (ADM--ES) end-to-end: (i) curated a compact, high-fidelity gold set; (ii) produced behaviour-aligned silver data via a RAG-conditioned mutator; (iii) assessed agent outputs with an Agent Judge along two orthogonal dimensions---Extraction Diagnostics (ED) and Behaviour Diagnostics (BD); and (iv) encoded Judge prescriptions into a durable recommendation map for reuse across iterations. In aggregate, the pipeline reframed evaluation from static scoring to an instrument for targeted and reproducible system control.

\subsection*{Principal findings.}
\begin{itemize}
  \item \emph{Behavioural transfer is feasible under adequate signal.} Across three of four primary contrasts, mutation redirected system behaviour toward expert style with statistically reliable and practically meaningful gains, strongest for Comment-level prescriptions (moderate for Gendered; large for Neurodiversity). Improvements tied to Expert-suggestion references were smaller and agent-dependent, consistent with lower information density.
  \item \emph{Extraction exhibits precision-biased, recall-limited performance.} The Neurodiversity agent maintained terminology and attribute fidelity when extracting, but systematically missed a non-trivial portion of expert evidence. Weighted ED placed performance mid-range with variance dominated by Correctness and Completeness.
  \item \emph{Judge validity is sufficient for steering, not adjudication.} BD facet scores were stable and aligned with expert intent. For ED, the Judge was reliable at the facet level but produced inconsistent on-prompt arithmetic, requiring external deterministic aggregation. Nevertheless, rationales and prescriptions were specific enough to drive prompt, decoding, and retrieval edits, which were organised into a UMAP-derived recommendation topology (7 ED clusters; 5 BD clusters).
\end{itemize}

\subsection*{Answers to the research questions.}
\begin{itemize}
  \item \textbf{RQ1 --- Feasibility of silver mutation.} Feasible and effective. Behaviour-aligned mutation yielded measurable gains in three of four tracks, with the largest effects where gold exemplars were richer and stylistically coherent, demonstrating that expert behaviour can be transferred via controlled context mutation without verbatim copying.
  \item \textbf{RQ2 --- Validity and utility of the Agent Judge.} Valid enough to steer. The BD facets tracked expert intent and produced actionable prescriptions; ED facetting was informative, but score aggregation must be externalised. The Judge is best viewed as a critic-coach that guides improvement rather than a final authority.
  \item \textbf{RQ3 --- Discovery of production failure modes.} Achieved. ED uncovered recall deficits (missed evidence; subtype/parent confusions) and pockets of over-extraction; BD revealed mild stylistic underfit (compressed register, attenuated hedging) despite strong semantic alignment. The recommendation map distilled these into reusable improvement blocks for successive releases.
\end{itemize}

\subsection*{Implications.}
The framework’s value lies less in a monolithic score than in a closed-loop process that (a) quantifies drift toward expert standards, (b) localises cognitive errors with facet-level granularity, and (c) codifies effective interventions as portable prescriptions. This enables teams to prioritise high-leverage changes---e.g., integrating mutated Comment prescriptions into synthesis and tightening ED evidence policies---while tracking attributable deltas over time.

\paragraph*{Final Thoughts.}
The proposed diagnostics convert agent evaluation from a retrospective audit into an engineered pathway for expert-behaviour transfer. In JobFair’s recruiter-assistant stack, the method produced reproducible evidence of behavioural cloning, surfaced extraction bottlenecks salient to users, and yielded a compact, navigable prescription map that productises expert oversight. The framework is thus a viable blueprint for diagnosing---and systematically improving---stochastic, tool-augmented LLM agents beyond this application.

\section{Comparative Positioning}

A central contribution of this work lies in positioning the proposed diagnostic framework relative to two established paradigms in LLM evaluation: \textit{LLM-as-a-Judge} and \textit{Agent-as-a-Judge}. Each of these approaches offers distinct advantages but also inherits limitations that our framework seeks to overcome.

\section*{LLM-as-a-Judge}
The LLM-as-a-Judge paradigm employs a single large model to score outputs against expert rubrics. This approach has demonstrated moderate reliability in settings such as summarization or dialogue evaluation, where surface-level coherence and factual sufficiency can be approximated through heuristic rubrics. Its strengths lie in scalability---thousands of outputs can be rapidly annotated---and in reproducibility, since evaluation depends only on fixed prompts and seeds. However, prior studies highlight two persistent weaknesses: (i) susceptibility to prompt bias and miscalibration, and (ii) limited visibility into multi-step reasoning failures \citep{gu2024survey_llm_as_a_judge}. A static Judge can identify that an output ``sounds'' less expert-like, but cannot attribute this deficiency to upstream breakdowns in evidence extraction, planning, or tool use. 

In contrast, our framework grounds evaluation in \textbf{curated golden datasets and silver mutations}, thereby anchoring judgments in explicit expert rationales. By operationalizing both Extraction Diagnostics (ED) and Behavior Diagnostics (BD), the framework moves beyond surface-level scoring and directly localizes failure modes, such as over-extraction or stylistic, behavior drift.

\section*{Agent-as-a-Judge}
A more recent extension, Agent-as-a-Judge, assigns evaluation to another agentic system that iteratively critiques and scores a target agent \citep{gu2024survey_llm_as_a_judge}. This approach improves on static LLM Judges by simulating task-oriented settings---capturing tool misuse, planning incoherence, or iterative refinement failures. Yet the recursive use of agents as evaluators introduces compounding stochasticity: meta-agents themselves hallucinate, diverge in reasoning style, and exhibit over-alignment to their own biases. Moreover, Agent-as-a-Judge studies often lack a stable ground truth, as they rely on emergent consensus across agent simulations rather than explicit expert annotations. 

Our framework addresses this fragility by hybridizing \textbf{agent-based evaluation with expert anchoring}: the Agent Judge is calibrated against golden annotations and silver-standard mutations, while its prescriptions are embedded into a vectorized recommendation map. This ensures that iterative critique is tethered to stable expert exemplars, reducing the risk of runaway bias reinforcement.

\section*{Positioning}
The proposed framework can be seen as a \textit{third path}: it retains the scalability and rubric-driven structure of LLM-as-a-Judge, while incorporating the dynamic, multi-step evaluation capacity of Agent-as-a-Judge. Crucially, it extends both by embedding prescriptions into a \textbf{Recommendation Map}, transforming evaluation from a terminal verdict into a mechanism for reusable knowledge transfer. Rather than scoring outputs in isolation, the framework constructs a navigable graph of expert-aligned improvement trajectories, enabling both diagnosis and steering. This not only addresses the stochasticity inherent in agentic LLMs but also establishes a reproducible protocol for expert behaviour transfer across tasks and domains.

\section*{Scope and Reproducibility}
\label{sec:scope-reproducibility}

The primary objective of this dissertation is to design and evaluate a diagnostic framework for detecting cognitive failures in multi-agent LLM systems using dynamic evaluation protocols and processing context mutation strategies.

This dissertation was conducted in close collaboration with industry. As a result, the empirical evaluation was intentionally scoped to the company’s production \emph{Gendered Language System (GLS)}, a proprietary multi-agent orchestration environment. The diagnostic framework was validated across two specialist expert agents within GLS, demonstrating that the proposed methods identify and steer cognitive failures consistently between distinct expert personas. While this constraint ensures relevance to a real, production-scale system where reliability has direct business impact, it also narrows external validity.

To establish broader generalisability, the diagnostic framework---which is designed to be architecture-agnostic and domain-independent---should be exercised on additional agentic frameworks and a wider taxonomy of tasks (e.g., alternative planning schemes, different tool-use policies, and varied application domains). Full coverage of the knowledge and behaviour space therefore requires follow-up studies on open-source agentic stacks and task families beyond GLS.

\paragraph*{Reproducibility.}
To support basic reproducibility, the complete experimental environment (code, dependencies, prompts, and configuration) has been containerised into a Docker image so that the pipelines can be executed on other machines with environment parity. Re-running the container reproduces the evaluation logic and metrics computation under fixed seeds where applicable. 

End-to-end replication of the reported results, however, depends on proprietary JobFair assets (including datasets and internal services) that cannot be distributed publicly. Researchers seeking to validate the results must obtain access credentials and data use approval from the JobFair management team. Without this approval, third parties can verify the build and execution of the diagnostic pipelines using the Docker image but cannot reproduce headline scores that rely on proprietary data.

\section{Future Work and Research Directions}
\label{sec:future-work}

This dissertation has introduced a novel \emph{diagnostic framework} for detecting \emph{cognitive failures in multi-agent LLM systems} and demonstrated its feasibility on the company's \emph{Gendered Language System (GLS)}. While the results confirm the methodology's potential, there are several areas for further development. This section consolidates future work into three main directions: \textbf{short-term enhancements}, \textbf{medium-term opportunities}, and a \textbf{long-term research vision}.

\subsection*{Short-Term Enhancements}
\label{sec:future-work-short}

The most immediate priority is the implementation of an \emph{Improvements Tracking} module, which was discussed in the methodology but remains underexplored. Such a component would enable continuous monitoring of agent performance over time, identifying patterns of improvement, stagnation, or degradation. By introducing temporal profiling and longitudinal drift detection, the framework could evolve from a \emph{static evaluation tool} into a \emph{dynamic monitoring system} capable of capturing changes in agent reliability after fine-tuning or system updates.

Another key short-term goal involves \emph{expanding experimentation beyond GLS}. While the corporate focus of this dissertation restricted testing to a single environment, future research should evaluate the framework on \emph{public benchmarks} such as \textbf{AgentBench} \citep{agentbench2024}, \textbf{MCPVerse} \citep{lei2025mcpverse}, and \textbf{GAIA}. These datasets would allow testing across diverse agentic workflows, coordination topologies, and failure patterns, ensuring the methodology's broader generalizability.

Finally, future work should enhance \emph{visualization and reporting capabilities}. Integrating \emph{UMAP-based cognitive failure maps}, recovery heatmaps, and interactive dashboards would make the framework more interpretable and usable in production-scale environments, particularly for developers and research teams overseeing complex multi-agent pipelines.

\subsection*{Medium-Term Research Opportunities}
\label{sec:future-work-medium}

A natural next step involves enabling \emph{adaptive diagnostics} through online learning techniques. Currently, intervention policies are static; incorporating reinforcement learning or multi-armed bandit approaches could allow the system to dynamically select the most effective context mutation strategies based on real-time performance feedback. This would improve both recovery rates and overall agent reliability.

Equally important is establishing \emph{cross-domain benchmarking}. Beyond GLS, multi-agent cognitive failures should be studied systematically across different orchestration frameworks. Building a shared repository of annotated failure cases---including hallucinations, coordination deadlocks, and planning breakdowns---would create a foundation for a standardized diagnostic benchmark, analogous to \textsc{MMLU} or \textsc{BIG-Bench}, but tailored to multi-agent evaluation. Such collaborative benchmarking would help define consistent metrics for measuring agent reliability and recovery efficiency across diverse setups.

\subsection*{Long-Term Research Vision}
\label{sec:future-work-long}

Looking further ahead, this work points toward the creation of \emph{self-diagnosing AI ecosystems}. By combining cognitive failure detection, dynamic evaluation, improvements tracking, and adaptive intervention strategies, future systems could autonomously monitor their own performance, adjust orchestration strategies, and minimize cascading errors without human oversight.

These capabilities would have significant implications for \emph{AI safety and reliability}, particularly as multi-agent LLM systems become embedded in \emph{high-stakes domains} like finance, healthcare, and robotics. The proposed framework could form the basis of \emph{audit-ready, interpretable evaluation pipelines}, supporting safe deployment in regulated environments.

Finally, an ambitious long-term goal is to establish an \emph{open-source diagnostic platform}. Such a platform would integrate dynamic evaluation protocols, standardized taxonomies of failure types, reusable cross-benchmark datasets, and advanced visualization tools. By enabling collaboration between \emph{academic researchers, industry practitioners, and open-source contributors}, it could accelerate innovation in agent diagnostics, fostering \emph{transparency, trust, and reliability} in large-scale multi-agent systems.

\begin{appendices}

\chapter{Prompts}
\label{appendix_introduction}

\lstdefinestyle{wrap}{
  basicstyle=\ttfamily\small,
  numbers=left, numbersep=6pt,
  breaklines=true, breakatwhitespace=false,
  columns=fullflexible, keepspaces=true,
  showstringspaces=false,
  frame=lines,
  xleftmargin=0pt, xrightmargin=0pt,
  postbreak=\mbox{\textcolor{gray}{$\hookrightarrow$}\space},
}

\appendix
\UseRawInputEncoding
\subsection*{Agent Mutator Prompt }
\label{ref:agent_mutator}
\begin{lstlisting}[style=wrap,,numbers=left,breaklines=true]
Your role is to help to mutate the job description to be more inclusive and accessible.

Context:
You are given:
1. Job Description – the original text of the posting.
2. Five example expert analyses – these show how an expert has identified and fixed gendered language or inclusivity issues in other job descriptions.
3. Target Recommendation – a recommendation generated by an automated system, which may need to be refined to match the expert’s style, logic, and behavior.

JOB DESCRIPTION:
${job_description}

EXAMPLE EXPERT ANALYSES:

Example 1:
${expert_examples_1}

Example 2:
${expert_examples_2}

Example 3:
${expert_examples_3}

Example 4:
${expert_examples_4}

Example 5:
${expert_examples_5}

TARGET RECOMMENDATION:
${target_recommendation}

Instructions:

1. Review the examples carefully to understand the expert’s style, reasoning, tone, and preferred phrasing patterns. Your aim to mutate text in Target Recommendation to match expert behavior. Pay attention to:
    * How they identify gendered_item
    * How they phrase the recommendation
    * How they construct the example fix
2. Evaluate the Target Recommendation:
    * If it is necessary, keep the original meaning but edit to match the expert’s style, logic, and behavioral patterns.
    * Ensure the gendered_item is completely unchanged, but the recommendation and example are modified to fit the expert’s style.
    * Maintain semantic alignment with the job description so that your suggestion is contextually correct.
3. When modifying:
    * Preserve the expert’s preferred tone and clarity level.
    * Apply the same balance between precision and accessibility used in the examples.
    * Adapt terminology and phrasing to fit the patterns observed in the examples.
\end{lstlisting}

\subsection*{BD Agent System Prompt }
\label{agent_bd_prompt}
\begin{lstlisting}[style=wrap,numbers=left,breaklines=true]
Role & Objective:

You are an Expert in LLM Agentic Systems and Agent Judge diagnosing why an agentic LLM’s OriginalSuggestion diverges from a MutatedSuggestion for a given JobDescription. Your job is to 

(1) pinpoint concrete divergences related to gendered or exclusionary language, tone, and content scope;
(2) infer the most probable sources of improvements (prompting, decoding, post-processing, tools, data coverage);
(3) propose actionable, testable interventions that will move system output toward the MutatedSuggestion

Evaluation Rubric (0–5 per facet; weighted sum → BDScore 0–100)

Score each facet on integers 0–5, then compute the weighted sum. 

Facet definitions and anchors:

1) tone_match (w=0.25): Degree to which the system sentence replicates the emotional, rhetorical, and formality tone of the expert sentence.
Anchors:
0 = tone inverted/inappropriate;
1–2 = major mismatch (e.g., overly casual or overly assertive);
3 = broadly aligned with some tonal drift;
4 = well-matched in stance and formality;
5 = indistinguishable tone match across emotional and rhetorical dimensions.

2) stylistic_fidelity (w=0.25): Similarity in surface-level features such as syntax, vocabulary register, and sentence rhythm.
Anchors:
0 = entirely different structure and register;
1–2 = significant simplification or mismatch in phrasing;
3 = generally similar, with minor vocabulary or rhythm shifts;
4 = close structural and lexical match;
5 = near-exact imitation in syntax, lexical choice, and flow.

3) manner_of_expression (w=0.25): Preservation of rhetorical and expressive style, including abstraction level, metaphor, and indirectness.
Anchors:
0 = opposite or incompatible expression strategy;
1–2 = major shift (e.g., metaphor removed or direct substituted for nuanced);
3 = partially retained expression with altered rhetorical technique;
4 = good retention of rhetorical device and abstraction style;
5 = full mirroring of expressive style, including devices, analogies, and abstraction level.

4) semantic_alignment (w=0.25): Accuracy in preserving the expert sentence's meaning, logical structure, and intent.
Anchors:
0 = meaning reversed or major distortion;
1–2 = partial preservation with omissions or added inferences;
3 = core meaning retained with minor shifts;
4 = meaning and logical relations intact;
5 = complete semantic fidelity with precise preservation of all content.

BD Score Calculation:

You can use code to calculate the BDScore

BDScore = 20 * (0.25 * tone_match + 0.25 * stylistic_fidelity + 0.25 * manner_of_expression + 0.25 * semantic_alignment)


Ground Rules

•	Base all findings on evidence from the four inputs; quote minimal spans as evidence.
•	Be specific and operational. 
•	Return JSON only matching the schema exactly.
•	Prioritize gender-neutrality and inclusivity aligned with the MutatedSuggestion, e.g., neutral pronouns, neutralized role nouns, removal of gender-coded adjectives but your aim to move system output toward the gold behavior by emitting similar patterns in language.
•	Your main aim is to transfer the behavioral, stylistic patterns from the MutatedSuggestion back to the OriginalSuggestion.

Reference Information:

•	OriginalSystemPrompt (string) - the original prompt which was used to create OriginalSuggestion

    ${OriginalSystemPrompt}

\end{lstlisting}

\subsection*{ED Agent System Prompt }
\label{ref:agent_bd_prompt}
\begin{lstlisting}[style=wrap,numbers=left,breaklines=true]
Role & Objective

You are an Expert in LLM Agentic Systems and Agent Judge. Given a JobDescription, an agentic LLM’s OriginalSuggestions, and the expert annotations (GoldenSuggestions) created by human expert, your task is to:

1.	Identify blind spots—recurrent failure patterns suggested in gendered_item.
2.	Attribute probable root causes (prompting, decoding, post-processing, scope/architecture, data/tools).
3.	Recommend targeted, testable remediations to push the system toward gold-like behavior with minimal risk.
4.	Propose counterfactual probes and dataset augmentations to confirm or refute each blind-spot hypothesis.

Rubrics:

**(1) Correctness** *(Weight: 0.35)*  
- Measures accuracy of OriginalSuggestions compared to the GoldenSuggestions.
- Checks if extracted entities, labels, and findings are exactly correct.
- Incorrect extractions reduce this score.

**(2) Completeness** *(Weight: 0.30)*  
- Measures recall: how many expert findings the system successfully captured.
- Missing diagnoses or missed clinical findings reduce this score.

**(3) Over-Extraction** *(Weight: 0.15)*  
- Penalizes extra or hallucinated findings not present in expert annotations (GoldenSuggestions).
- More extra findings = stronger penalty.

**(4) Detail Accuracy** *(Weight: 0.10)*  
- Measures precision in associated attributes: severity, stage, onset, timing, test values, locations, etc.
- Even if diagnoses match, wrong attributes reduce this score.

**(5) Terminology Consistency** *(Weight: 0.05)*  
- Ensures that the terms used match expert labels or are exact synonyms.
- Wrong or ambiguous terms reduce this score.

**(6) Reasoning Alignment** *(Weight: 0.05)*  
- Evaluates whether the OriginalSuggestions follows the same diagnostic reasoning process as the GoldenSuggestions.
- A correct answer with wrong reasoning gets partial credit.

Scoring Rules

You can use code to calculate the scores.

Each dimension receives a **sub-score from 0.0 to 1.0** using the following interpretation:

- **1.0** → Perfect alignment with expert annotations.
- **0.8–0.9** → Minor, clinically insignificant deviations.
- **0.6–0.7** → Several mismatches, but still partially correct.
- **0.4–0.5** → Many errors; unreliable for diagnostics.
- **<0.4** → Unusable; almost completely wrong.

Use the following formula for the **Final Score**:

Final Score = 10 * (  
    0.35 * Correctness +  
    0.30 * Completeness +  
    0.15 * (1 - OverExtraction) +  
    0.10 * DetailAccuracy +  
    0.05 * TerminologyConsistency +  
    0.05 * ReasoningAlignment  
)

**Final Score Scale**:
- **9.0 – 10** → Expert-level, perfect extraction.
- **8.0 – 8.9** → High-quality, clinically acceptable.
- **6.5 – 7.9** → Moderate quality, requires review.
- **4.0 – 6.4** → Low quality, unreliable extraction.
- **1.0 – 3.9** → Failed; cannot be trusted.

Step-by-Step Evaluation Process:

**Step 1. Identify Matches, Misses, and Extras**  
- **True Positives (TP)**: Findings present in both expert & system.
- **False Negatives (FN)**: Findings in expert but missing in system.
- **False Positives (FP)**: Findings in system but absent in expert.

**Step 2. Score Each Dimension**  
- Correctness = TP / (TP + FP)  
- Completeness = TP / (TP + FN)  
- Over-Extraction = FP / (TP + FP)  
- Detail Accuracy = Percentage of correct attributes matched.  
- Terminology Consistency = Ontology similarity score.  
- Reasoning Alignment = Semantic similarity of explanations.

**Step 3. Calculate Final Score**  
- Apply the weighted formula.
- Round to **two decimal places**.


Ground Rules
•	Be evidence-driven: quote minimal spans from all inputs.
•	if GoldenSuggestions doesn't contain any or some elements but OriginalSuggestions does, it indicates a potential blind spot in the system's understanding or extraction capabilities for OriginalSuggestions.
•	Generalize cautiously from the instance to a hypothesized pattern; state the trigger conditions.
•	Do not rewrite the gold or produce free-form prose in the final answer; return JSON only per the schema.


Reference Information:

•	OriginalSystemPrompt (string) - the original prompt which was used to create OriginalSuggestion

    ${OriginalSystemPrompt}
\end{lstlisting}

\end{appendices}

\clearpage 

\nocite{*}
\printbibliography

\end{document}